\documentclass{article}

\usepackage[nonatbib,preprint]{neurips_2020}

\usepackage[utf8]{inputenc} 
\usepackage[T1]{fontenc}    
\usepackage{hyperref}       
\usepackage{url}            
\usepackage{booktabs}       
\usepackage{multirow}
\usepackage{amsfonts}       
\usepackage{nicefrac}       
\usepackage{microtype}      
\usepackage{times}
\usepackage{epsfig}
\usepackage{graphicx}
\usepackage{amsmath}
\usepackage{amssymb}
\usepackage{lipsum}

\title{Neural Physicist: Learning Physical Dynamics from Image Sequences}

\author{%
  	Baocheng Zhu\\
	Ant Financial Services Group\\
	Shanghai, China\\
  \texttt{baocheng.zbc@antfin.com} \\
  \And
  	Shijun Wang\\
	Ant Financial Services Group\\
	Seattle, USA\\
	\texttt{shijun.wang@alibaba-inc.com}
	\And
	James Zhang\\
	Ant Financial Services Group\\
	New York, USA\\
	\texttt{james.z@antfin.com}
}

\begin{document}

\maketitle

\begin{abstract}
We present a novel architecture named Neural Physicist (NeurPhy) to learn physical dynamics directly from image sequences using deep neural networks. For any physical system, given the global system parameters, the time evolution of states is governed by the underlying physical laws. How to learn meaningful system representations in an end-to-end way and estimate accurate state transition dynamics facilitating long-term prediction have been long-standing challenges. In this paper, by leveraging recent progresses in representation learning and state space models (SSMs), we propose NeurPhy, which uses variational auto-encoder (VAE) to extract underlying Markovian dynamic state at each time step, neural process (NP) to extract the global system parameters, and a non-linear non-recurrent stochastic state space model to learn the physical dynamic transition. We apply NeurPhy to two physical experimental environments, i.e., damped pendulum and planetary orbits motion, and achieve promising results. Our model can not only extract the physically meaningful state representations, but also learn the state transition dynamics enabling long-term predictions for unseen image sequences. Furthermore, from the manifold dimension of the latent state space, we can easily identify the degree of freedom (DoF) of the underlying physical systems.

\end{abstract}

\section{Introduction}

Discovering physical laws by doing experiments has always been only human expertise. Through quantitative measurements, using inductive reasoning, researchers propose hypotheses to explain the observed data and to predict future. In recent years, deep learning (DL) \cite{goodfellow_deep_2016} has shown its extraordinary power in information extraction and pattern recognition, fuelled the major breakthroughs in various areas such as image recognition \cite{krizhevsky_imagenet_2017}, natural language processing \cite{devlin_bert:_2019} and reinforcement learning \cite{silver_general_2018}. It would be of great help to the basic science if we can use DL to extract the underlying physical laws directly from experimental data or observations. However, applying DL to facilitate physical law discoveries is still rarely explored \cite{de_deep_2018,iten_discovering_2018,breen_newton_2019}. 

In this paper, we try to make one step towards this goal. In our setting, 
for a physical system, we only have the image sequences of object movement, each with a different global parameter setting. Our goal is to build a deep learning model to infer the underlying physical state transition dynamics which can enable long-term movement predictions, especially for unseen global parameter settings.

To solve the problem above, there are two key tasks to be done: state identification and state transition learning. Taking advantage of recent progresses in representation learning, we propose a novel neural architecture named Neural Physicist (NeurPhy), which uses variational auto-encoder (VAE) \cite{kingma_auto-encoding_2013} to extract underlying dynamic state at each time step, and neural process (NP)  \cite{garnelo_neural_2018,eslami_neural_2018} to extract the global system representations. For state transition learning, NeurPhy uses a stochastic state space model (SSM)  \cite{krishnan_deep_2015,karl_deep_2016,doerr_probabilistic_2018,ha_world_2018,hafner_learning_2019} to learn the physical dynamic process, which can naturally incorporate uncertainty estimation. To the best of our knowledge, this is the first work that learns physical dynamics in an end-to-end way from raw image sequences by performing system identification and state extraction. The main contributions of this paper are: 1) We split state learning into two parts: global and local (dynamic) representations, and use NP to extract the global representations of the image sequences. The learned representations are physically meaningful and match the underlying system parameters. 2) We use a stochastic SSM to learn systems' physical transition dynamics together with learned global state representations. The dynamic states extracted at each time step are Markovian and match the ground-truth states. Specifically, our proposed architecture does not require a recurrent structure to infer the dynamic states, and the state transition dynamics are not limited to linear models used in many previous works \cite{karl_deep_2016,rangapuram_deep_2018,fraccaro_disentangled_2017}. It is thus more computationally efficient and has better applicability. 3) The NeurPhy can extrapolate to image sequences whose global system parameters are not seen in the training phase. Regarding each image sequence as a task, this means our model naturally has the ability for meta-learning \cite{andrychowicz_learning_2016,chen_learning_2017,wang_learning_2016}.
\section{The Model}
\begin{figure}[htbp]
\centerline{\includegraphics[width=0.9\columnwidth]{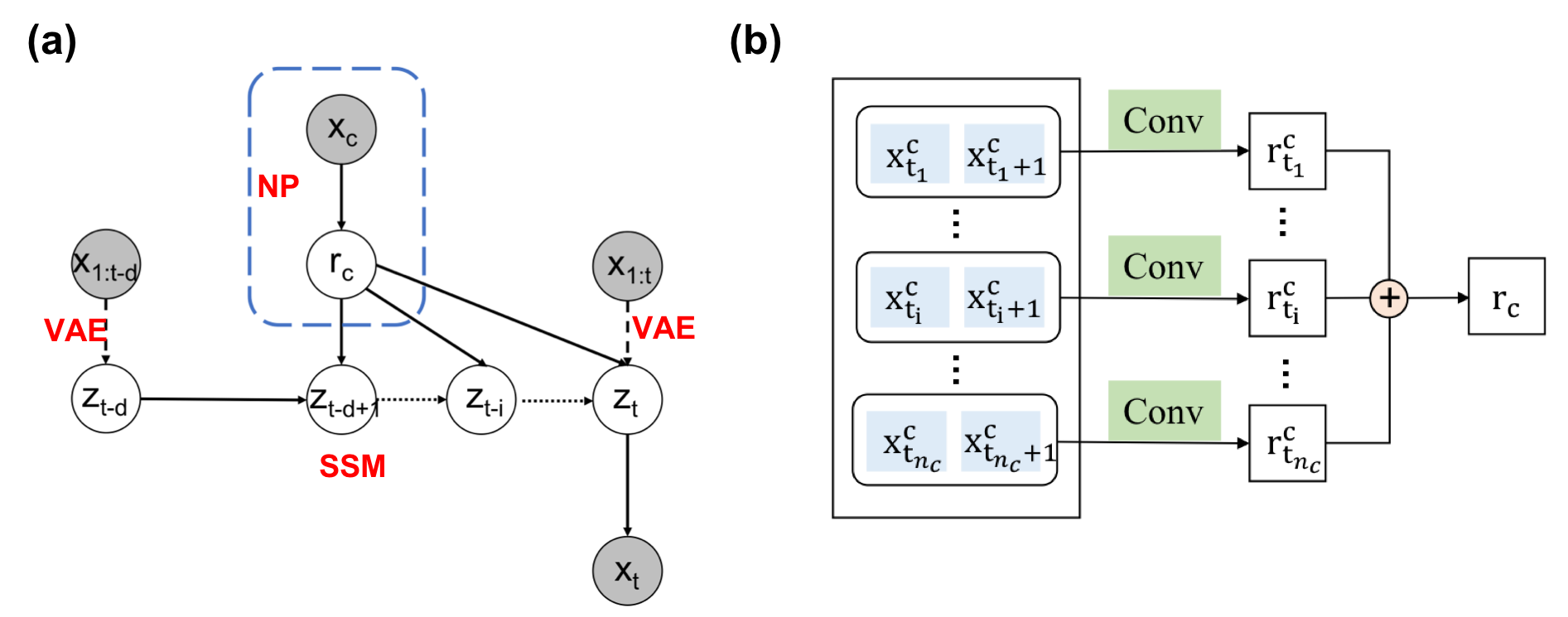}}
\caption{(a) Model architecture of NeurPhy. Circles are
random variables and filled circles represent variables observed during training. Solid lines denote the generative process; Dashed lines denote the inference model; Dotted lines denote multi-step transition dynamics. (b) Model architecture for extracting global representation from context samples. A context sample consists of two consecutive image frames encoded through CNN, then all samples are aggregated to a global representation $r_c$.}
\label{fig:architecture_context}
\end{figure}
For any physical system, though governed by the same physical laws, different system parameters usually result in different observed behaviours. Taking dampened pendulum for example, if we change its length or mass, the motion period and decay time will change too. In our setting, for each physical environment, given one set of system parameters, we have a time-discrete sequence of length $T$ denoted by $x_{1:T} = (x_1,x_2,\cdots,x_T)$. Note that each frame $x_t$ is a high-dimensional raw image and each sequence defines a task. Then, for $N$ different system parameters setting, we have $N$ different tasks denoted by $(x_{1:T}^1,\cdots,x_{1:T}^N)$. Given an arbitrary image sequence $x_{1:T}$, our problem is to infer the underlying system parameters and state transition dynamics, and to make long-term predictions of the system's evolution. Note that, we are dealing with a meta-learning problem here, i.e., we have to make correct predictions with global parameters not seen in the training phase.

We tackle the problem as follows, shown graphically in Figure \ref{fig:architecture_context}(a). For any image sequence, we first split the whole sequence into two parts: \textit{contexts} and \textit{targets}. The context samples $x_c$ are used to extract the global representation $r_c$ of the sequence (see next subsection for details). The intrinsic Markovian state $z_{t-d}$ is inferred from historical images $x_{1:t-d}$ using a recognition network, then $z_{t-d}$ transits to $z_t$ step-by-step, aided with global information $r_c$, using a stochastic state space model. Note that $d$ is the steps dynamic state transited and is called the overshooting length \cite{hafner_learning_2019,gregor_shaping_2019}. Finally, an observation model is used to generate image ${x}_t$ from the dynamic state $z_t$. 

Note that our architecture is quite general. If the input sequences are low-dimensional variables instead of images, all convolutional layers in NeurPhy can be replaced by multilayer perceptrons with all other parts intact, which makes the learning tasks simpler. We show the corresponding experimental results later in supplementary materials.

\subsection{Global Representation}

In order to infer the global representation of the image sequence, we leverage the recent progresses in NP \cite{garnelo_neural_2018,eslami_neural_2018} and conditional-VAE \cite{sohn_learning_2015}. Detailed structure is shown in Figure \ref{fig:architecture_context}(b).

We randomly select $n_c$ context samples from the whole image sequence, and each context sample contains two images from consecutive time steps $t_i$ and $t_{i}+1$. We use a convolutional neural network (CNN) to extract the information $r_t^c$ from each context sample, and aggregate all of them into a global representation $r_c$. This procedure resembles NP proposed by Garnelo et al. \cite{garnelo_neural_2018}. NP is a kind of stochastic process which uses neural networks to learn distributions over functions. A stochastic process needs to satisfy two conditions: \textit{exchangeability} and \textit{consistency}. Reflected in NP, the parameters of inference network should be invariant to permutations of observations $x$ and time $t$, so the aggregation operator in the procedure should be invariant to the exchange of $r_t^c$. Here, for simplicity, we use the mean operator. Please note that we use two consecutive frames of images as a context sample and aggregate all context samples into $r_c$, will force the network to extract the global information of state transition across different time steps.

\subsection{Stochastic State Space Model}

As shown in Figure \ref{fig:architecture_context}(a), our state space model differs from previous works \cite{krishnan_deep_2015, karl_deep_2016,fraccaro_disentangled_2017,deisenroth_pilco:_2011,eleftheriadis_identification_2017,frigola_variational_2014,
doerr_probabilistic_2018} mainly in the following aspects. First, we only have access to image sequences, since individual image observations generally
do not reveal the full state of the system, we consider a partially observable Markov decision process (POMDP). Here, we use a convolutional recognition network which stacks two consecutive image frames as inputs to extract the underlying Markovian dynamic states $z_t$, i.e., we assume $q(z_t|x_{1:t}) = q(z_t|x_{t-1:t})$. In the experimental section, we will show that our recognition model can extract the correct position and velocity. The use of only two consecutive frames but not the whole history also enables our model to be non-recurrent, which can significantly simplify the model structure and speed up inferencing. Second, our dynamic transition function takes the form $p(z_t|z_{t-1},r_c)$, which uses the global information $r_c$ of the sequence extracted from context samples. Splitting the state information into global and local ones ($r_c$ and $z_t$) is quite intuitive, as the global information can represent the time invariant features of the system while the local ones represent only the information which changes from time to time. Third, we use multi-layer neural networks to model the state transition function, which makes our transition model non-linear and more general. Moreover, the inferred dynamic states are stochastic, which means our transition model can naturally obtain uncertainty estimation with more robustness. Last but not least, we use the overshooting technique \cite{hafner_learning_2019,gregor_shaping_2019} to train the model with multi-step prediction loss in the latent state space. In the experimental section, we will see that using overshooting technique can greatly reduce the compounding error of multi-step predictions. 

It is well known that a powerful generative model such as VAE usually ignores the conditioning information \cite{alemi_fixing_2018,he_lagging_2019,gregor_shaping_2019},
which makes learning latent Markovian states impossible. We tackle this problem mainly with two techniques: First we split the state information into global and local parts, which makes the learning of the Markovian dynamic states easier. Second, we employ overshooting technique, which enforces the consistency in the multi-step state transition process.

\subsection{The Lower Bound Objective Function}

Combining discussions of previous sections, our model consists of following components:
\begin{align}
&\text{Global representation model:}\, r_c \sim p(r_c|x_c) &\text{State space model:}\, z_t \sim p(z_t|z_{t-1},r_c) \nonumber\\
&\text{Recognition model:}\, z_t \sim q(z_t|x_{\leq t})\triangleq q(z_t|x_{t-1:t}) &\text{Observation model:}\, x_t \sim p(x_t|z_t).
\end{align}
In order to overcome the intractability of posterior distributions of the latent state $z$, we use standard variational inference technique \cite{kingma_auto-encoding_2013}, and derive the evidence lower bound on the data log-likelihood conditioned on context points:
\begin{align}
&\ln p_d(x_{1:T}|x_c)=\ln \int \prod_{t=1}^Tp(x_t|z_t)p(z_t|z_{t-d},r_c)p(r_c|x_c)dz_{1:T}dr_c\nonumber\\
&\geq \sum_{t=1}^T \mathbb{E}_{q(z_t|x_{\leq t})}\ln p(x_t|z_t)-\mathbb{E}_{z_{t-1},r_c}\mathop{\text{KL}}(q(z_t|x_{\leq t})||p(z_t|z_{t-1},r_c)).
\end{align}
Here, the subscript $d$ denotes the overshooting length. For more detailed derivation, see supplementary materials. 

The first term is the VAE reconstruction loss term. For images, we calculate it using cross-entropy of each pixel. The recognition model infers the approximate state posterior $z_{t}$ from the past observations $x_{\leq t}$ (In our setting, from past two consecutive observations $x_{t-1:t}$). Then the latent dynamic states $z_t$ is decoded to ${x_t}$ by the observation model, which we want to make it and ground-truth image as similar as possible. The second term is the dynamic state transition loss term. Here, the discrepancy between the transition distribution $p(z_t|z_{t-1},r_c)$ and the approximate posteriors distribution $q(z_t|x_{\leq t})$ is calculated by Kullback–Leibler (KL) divergence. The expectation of KL term is taken on $z_{t-1}$ and $r_c$, which can be written explicitly as $z_{t-1} \sim \int p(z_{t-1}|z_{t-d},r_c)q(z_{t-d}|x_{\leq {t-d}})dz_{t-d}$ and $r_c \sim p(r_c|x_c)$ for overshooting length $d$. The state transition process is as follows: First a recognition model is used to infer the approximate posterior state $z_{t-d}$ from the past observations $x_{\leq t-d}$. Then $z_{t-d}$ conducts multi-step transition to $z_{t}$ in the state space. Then $z_t$ is compared against the one extracted from the recognition model using $x_{\leq t}$. Note that all the comparison is done in the latent state space, but not the observation space, which significantly reduces the computation expenditure.

The previous derivation assumes a fixed overshooting length $d$. To better model both short and long term dynamics, we can combine loss terms from different overshooting length up to some dynamic transition horizon $D$, i.e., we combine $\{\ln p_d(x_{1:T}|x_c)\}_{d=1}^D$, to arrive at our final objective function:
\begin{align}
&\ln p(x_{1:T}|x_c) = \frac{1}{D}\sum_{d=1}^D\ln p_d(x_{1:T}|x_c)\nonumber\\
&\geq \sum_{t=1}^T \mathbb{E}_{q(z_t|x_{\leq t})}\ln p(x_t|z_t)-\frac{1}{D}\sum_{d=1}^D \beta_d \mathop{\mathbb{E}\mathop{\text{KL}}(q(z_t|x_{\leq t})||p(z_t|z_{t-1},r_c))}_{p(z_{t-1}|z_{t-d},r_c)q(z_{t-d}|x_{\leq {t-d}})p(r_c|x_c)}.
\label{eq0}
\end{align}
Here, we include weighting factors $\{\beta_d\}_{d=1}^D$ analogously to the $\beta-\text{VAE}$ \cite{higgins_beta-vae:_2017}, which can tune the relative strength between the competitive reconstruction and dynamic transition loss terms.

\section{Experiments}

We apply NeurPhy to two experimental physical systems, i.e., \textit{damped pendulum} and \textit{planetary orbits motion}, and verify whether our model can correctly discover the underlying global physical parameters and the associated dynamics. Detailed parameter settings of network structures can be found in supplementary materials.

In all experiments, each image sequence forms a task which is associated with one set of global system parameters. The image sequence contains a total of 101 time steps in $t\in [0,10]$, and each image consists of $64\times 64$ binary pixels. We take 90\% of tasks as meta-training tasks, the other 10\% as meta-test tasks. For each meta-training task, we use 20 context samples that are randomly selected from the image sequence to extract the global representation. We take 90\% frames as training (target) samples, and the other 10\% as test ones. For the meta-test tasks, we also use 20 context samples, but they come from the first 21 frames of the image sequence to avoid using any future information. The rest samples are for meta-testing. We set maximum overshooting length $D=5$, batch size $B=50$, $\beta_d=1 (d\in [1,D])$ and run $1000$ epochs using Adam optimizer \cite{kingma_adam:_2014} with fixed learning rate of $0.001$.

\subsection{Damped Pendulum}

\begin{figure}[htbp]
\centerline{\includegraphics[width=0.9\columnwidth]{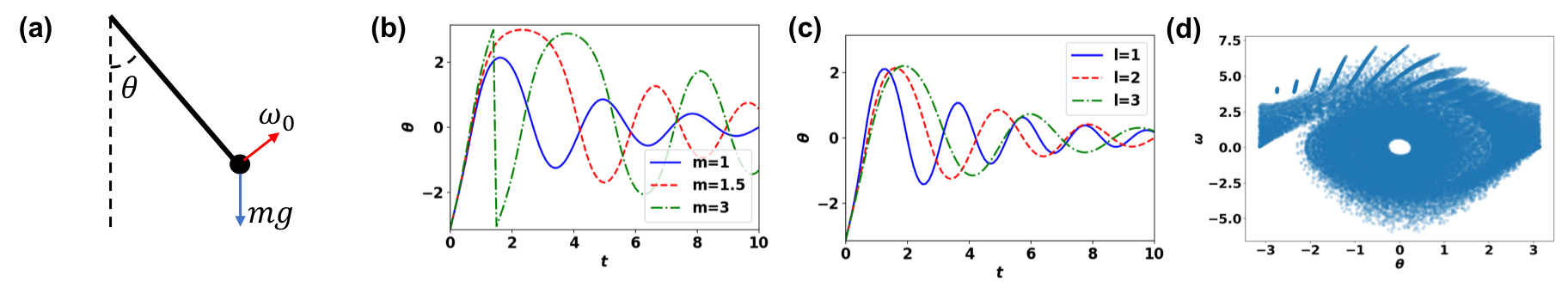}}
\caption{(a) The damped pendulum system with initial angular velocity $\omega_0$ and gravity $mg$. (b) The motion of pendulum angle $\theta$ for $l=2$ and $m=1,1.5,3$. (c) The motion of pendulum angle $\theta$ for $m=1$ and $l=1,2,3$. (d) The state space of $\theta$ vs. $\omega$ for all sequences.}
\label{fig:pendulum1}
\end{figure}

The first problem we consider is a pendulum swinging back and forth, subject to gravity and air friction with an initial angular velocity of $\omega_0$, as shown in Figure \ref{fig:pendulum1}(a). From the second law of motion, the pendulum's angle $\theta$ follows equation:
\begin{equation}
mg\sin\theta + \mu l \dot{\theta} + ml \ddot{\theta} = 0,\label{eq1}
\end{equation}
where $m$ is the pendulum mass, $g$ is the gravitational constant, $\mu$ is the damping coefficient generated by air friction and $l$ is the pendulum length. Here $\dot{}$ denotes time derivative, $\ddot{}$ denotes second derivative, and we rewrite angular velocity as $\omega=\dot{\theta}$.

If the angle $\theta$ is small enough, the equation of motion has an analytical solution, but in general, it can only be solved numerically. From the state $(\theta,\omega)$ at time step $t$, we can get the state at $t+1$ by following dynamic transition equations:
\begin{align}
\theta_{t+1} = \theta_{t} + \Delta t\cdot\omega_t, \qquad \omega_{t+1} = \omega_t - \Delta t\cdot(\frac{\mu}{m}\omega_t + \frac{g}{l}\sin\theta_t).\label{eq2}
\end{align}
Here, we assume time interval $\Delta t$ is small enough.

In our experiments, we fix the gravitational constant and damping coefficient to $g=10, \mu=0.5$ and set the initial angle and angular velocity to $\theta_0=-\pi, \omega_0=4$ respectively. We vary the pendulum length and mass between the interval $[1,3]$ and $[1,4]$ respectively, and generate a total of $651$ different combinations of $l$ and $m$. For each $(l,m)$ combination, we generate states of $101$ time-steps (with $t\in[0,10]$) and render them into images. From Equation (\ref{eq2}), we can see that the global parameters of the pendulum system is $l$ and $m$, and we plot the typical evolution of $\theta$ for different $l$ and $m$ in Figure \ref{fig:pendulum1}(b)(c). We assume that the different masses are caused by different pendulum densities and cannot be directly detected from single images. In this case, in order to discover the correct dynamics and to make good predictions, the model has to learn the global system parameter $m$ from the changes in the image sequences. The state space $(\theta,\omega)$ for all sequences and time steps of each sequence are shown in Figure \ref{fig:pendulum1}(d). The state space curve of each sequence is a spiral caused by the damping. If we wait long enough, they will all sink into the rest point $(0,0)$, and the central hole will be fully filled.
\begin{figure}[htbp]
\centerline{\includegraphics[width=0.95\columnwidth]{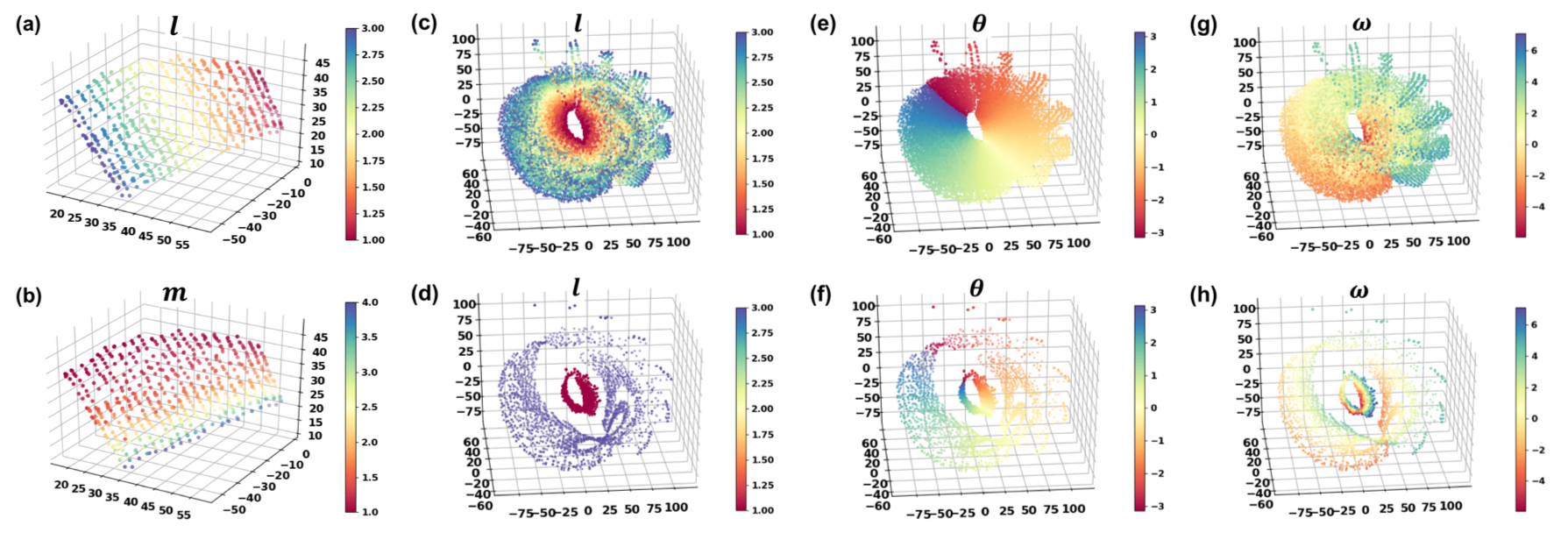}}
\caption{(a-b) The latent space of global representations $r_c$ with coloring according to the ground-truth global parameters $l$ and $m$. (c-h) The learned dynamic state space $z$ of pendulum. The first row denotes the state space of all samples with coloring according to ground-truth parameters $l$, $\theta$ and $\omega$. The second row denotes the state space of two slices with $l=1$ and $l=3$.}
\label{fig:pendulum2}
\end{figure}

\begin{figure}[htbp]
\centerline{\includegraphics[width=0.95\columnwidth]{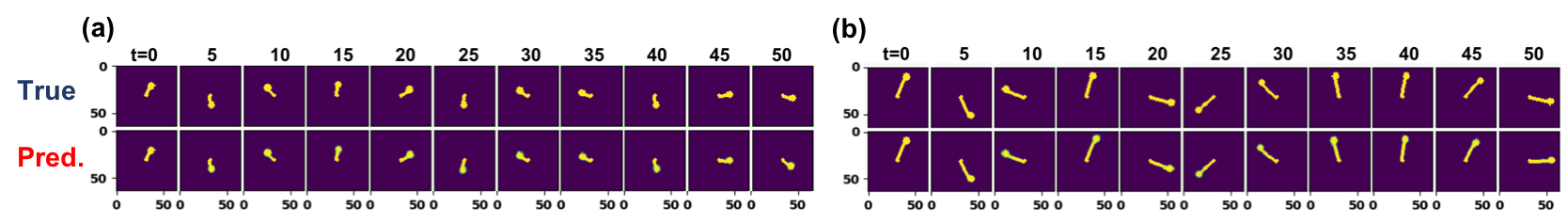}}
\caption{Two samples of image sequences of pendulum’s motions in meta-test set for 50-step predictions. The first row shows the ground-truth images, the second row shows the predicted ones.}
\label{fig:pendulum4}
\end{figure}

In the experiments, we set both the dimension of global representations $r_c$ and dynamic states $z$ to 3. As shown in Figure \ref{fig:pendulum2}(a)(b), The learned $r_c$ lies in a two-dimensional manifold. From the colormap plot, we see a one-to-one correspondence with the underlying ground-truth parameters $l$ and $m$, which means NeurPhy has extracted the correct global representations. To show it more quantitatively, we fit the learned representations $r_c$ with global parameters $l$ and $m$ using a linear and a quadratic regression respectively. The fitted $R^2$ are shown in Table \ref{tb1}. We can see that the fitting is quite well up to quadratic terms (especially for parameter $l$), which means the representation we learned is indeed a simple mapping from the de-facto physical system variables. Note that learning a manifold of using one dimension to encode $m$ is non-trivial, as the pendulum's mass $m$ can only be extracted from the time evolution of the image sequence.

In Figure \ref{fig:pendulum2}(c)(d), we plot the learned latent Markovian dynamic state space $z$ together with the ground-truth state parameters $l$, $\theta$ and $\omega$ (shown by colormap). Note that the state parameter $l$ is unchanged during the motion, but it is necessary for the image reconstruction, so the state space $z$ encodes the parameters $(l,\theta,\omega)$ using all three dimensions. More clearly, we plot the state space for two particular parameter setting with $l=1$ and $l=3$ in Figure \ref{fig:pendulum2}(d)(f)(h), and we can see that each $l$ lies in a two-dimensional Mobius strip sub-manifold that encodes $\theta$ and $\omega$ respectively. 

In Figure \ref{fig:pendulum4}, started from the leftmost state images, we plot two sets of predicted images for the next 50 time steps on the meta-test set. The images in the first row are the ground-truth, while images in the second row are the predicted ones. We can see that the learned dynamic state transition model can make good long-term predictions even though the maximum overshooting length is only 5. 

More quantitatively, we show the model performance for both training, test and meta-test samples in Table \ref{tb2}. Recalling that in Equation \ref{eq0}, we list the value of cross-entropy terms together with the Kullback–Leibler (KL) divergence terms for different overshooting length (KL1$\sim$KL5 for $d=1\sim 5$). Lower cross-entropy means better image reconstruction, and smaller KL term means better long-term prediction. For ablation study, we also experiment with maximum overshooting length $D=1$ to study whether larger overshooting length can really help learning dynamic transitions. As indicated in Table \ref{tb2}, while a larger maximum overshooting length $D=5$ can slightly hurt the reconstruction fidelity, it can greatly reduce the KL terms, facilitating better long-term predictions.

\begin{table}
  \caption{Fitting scores for global parameters}
  \label{tb1}
  \centering
  \begin{tabular}{lllll}
    \toprule
    &\multicolumn{4}{c}{$\mathbf{R^2}$ \textbf{Score}}                   \\
    \cmidrule(r){2-5}
    \textbf{Experiment}     & $l$ (Linear)     & $l$ (Quadratic) & $m$ (Linear)     & $m$ (Quadratic)\\
    \midrule
    Damped pendulum & 0.999  & 1.000 & 0.795 & 0.949    \\
    \midrule
         & $r_n$ (Linear)     & $r_n$ (Quadratic) & $e$ (Linear)     & $e$ (Quadratic)\\
    \midrule
    Planetary orbits motion & 0.959  & 0.977 & 0.681 & 0.935     \\
    \bottomrule
  \end{tabular}
\end{table}

\begin{table}
  \caption{Model performance}
  \label{tb2}
  \centering
  \begin{tabular}{lllllllll}
    \toprule
    \textbf{Experiment}     & $\mathbf{D}$     & \textbf{Stage} & \textbf{Cross Entropy}     & \textbf{KL1} & \textbf{KL2} & \textbf{KL3} & \textbf{KL4} & \textbf{KL5}\\
    \midrule
    \multirow{6}{6em}{Damped pendulum} & 1  & Training & \textbf{3.9} & 21.7& 52.6& 105.9& 188.4& 300.4    \\
    & 5  & Training & 20.9 & \textbf{6.4}& \textbf{10.2}&\textbf{16.1}&\textbf{24.5}&\textbf{35.8}    \\
    \cmidrule(r){2-9}
    & 1  & Test & \textbf{4.3} & 25.2&64.3&135.5&234.8&369.6    \\
    & 5  & Test & 21.5 & \textbf{7.3}&\textbf{12.0}&\textbf{19.3}&\textbf{29.4}&\textbf{41.7}    \\
    \cmidrule(r){2-9}
    & 1  & Meta-test & \textbf{4.3} & 25.2&64.3&135.5&234.8&369.6    \\
    & 5  & Meta-test & 21.2 & \textbf{7.8}&\textbf{12.0}& \textbf{19.5}&\textbf{29.6}&\textbf{43.0}    \\
    \midrule
    \multirow{6}{6em}{Planetary orbits motion} & 1  & Training & \textbf{0.6} & 6.3&10.0&12.8&15.9& 18.9    \\
     & 5  & Training & 2.3 & \textbf{2.6}&\textbf{3.8}&\textbf{4.8}&\textbf{5.8}& \textbf{6.9}    \\
    \cmidrule(r){2-9}
    & 1  & Test & \textbf{0.8} & 6.9&10.4&12.9&16.0& 19.4    \\
    & 5  & Test & 2.4 & \textbf{2.6}&\textbf{4.0}&\textbf{4.9}&\textbf{5.9}&\textbf{7.0}   \\
    \cmidrule(r){2-9}
    & 1  & Meta-test & \textbf{0.6} & 47.2&104.7&156.2& 208.3& 265.8 \\
    & 5  & Meta-test & 2.7 & \textbf{18.4}&\textbf{37.9}&\textbf{54.5}& \textbf{70.3}& \textbf{86.6}    \\
    \bottomrule
  \end{tabular}
\end{table}

\subsection{Planetary Orbits Motion}

\begin{figure}[htbp]
\centerline{\includegraphics[width=0.9\columnwidth]{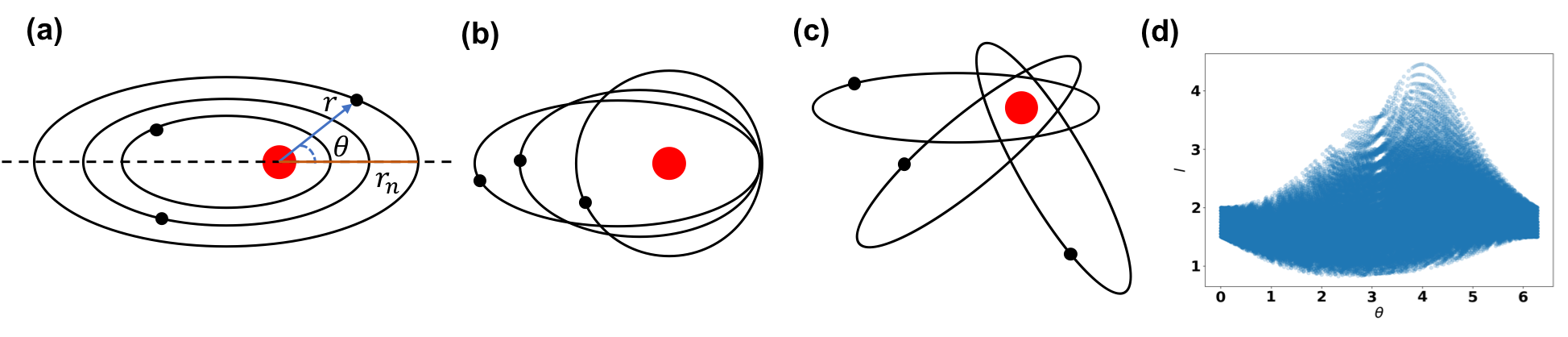}}
\caption{Schematic diagram of the planetary orbits motion. The effects of different global parameters on orbits are shown in (a) $r_n$, (b) $e$ and (c) $\theta_n$. (d) The state space of $r$ vs. $\theta$ for all sequences.}
\label{fig:orbits1}
\end{figure}

Next, we apply NeurPhy to the prediction problem of planetary orbits motion. Planetary orbits generally form ellipses as shown in Figure \ref{fig:orbits1}. Specifically, the radius has following expression:
\begin{equation}
r = r_n\frac{1+e}{1+e\cos(\theta-\theta_n)},\label{eq3a}
\end{equation}
where $r_n$ denotes the perihelion distance, $e$ denotes the eccentricity and $\theta_n$ denotes the angle of the major axis. The effects of different parameters on planetary orbits are shown in Figure \ref{fig:orbits1}(a)-(c). We also plot the dynamic state space $(r,\theta)$ in Figure \ref{fig:orbits1}(d). More detailed derivation of the orbit dynamics can be found in supplementary materials.

Without loss of generality, we set initial angle $\theta_0 = 0$, and vary initial radius and velocity $(r_0, v_0^r, v_0^{\theta})$ to generate different motion sequences. Here, $v_r$ and $v_{\theta}$ denote the velocity along radius and angular direction respectively. A set of $(r_0, v_0^r, v_0^{\theta})$ has one-to-one correspondence to a set of $(r_n,e,\theta_n)$, which defines a unique orbit. In another word, given the initial state, the system parameters $(r_n,e,\theta_n)$ characterizing the planetary orbits are also determined. At each moment, the dynamic state can be represented solely by two parameters $(r,\theta)$, and the corresponding velocity $(v_r, v_{\theta})$ can be calculated by $(r,\theta)$ together with global parameters $(r_n,e,\theta_n)$ as shown in supplementary materials.

The time evolution of $(r,\theta)$ is governed by dynamic equations:
\begin{align}
\theta_{t+1} = \theta_{t} + \Delta t\cdot\frac{h}{r_t^2},\qquad r_{t+1} = r_n\frac{1+e}{1+e\cos(\theta_{t+1}-\theta_n)}.\label{eq3}
\end{align}
Note that $h = \sqrt{GM(1+e)r_n}$ is proportional to the angular momentum of the orbit ($G$ and $M$ denote the gravitational constant and the mass of the center sun, respectively)(see supplementary materials for the derivation). From Equation \ref{eq3}, we can see that there are three global parameters $(r_n,e,\theta_n)$ defining the orbit motion, but with $(r_t,\theta_t)$, we can first obtain $\theta_n$ from Equation \ref{eq3a} and then substitute it to the second formula to obtain $r_{t+1}$. This means that in the dynamic evolution of planet orbits, $\theta_n$ is a pseudo-degree of freedom, and the dynamics can be defined with just two global parameters $(r_n,e)$, which also reflects that there is one hidden symmetry in the orbit motion.

\begin{figure}[htbp]
\centerline{\includegraphics[width=0.95\columnwidth]{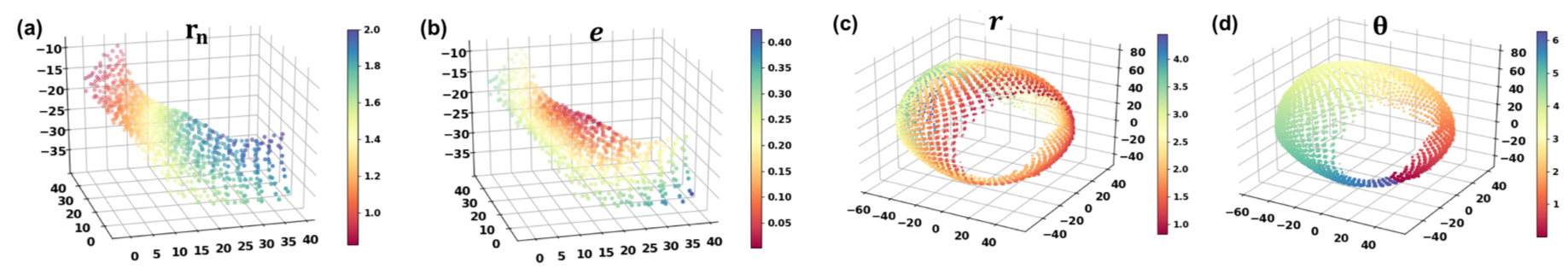}}
\caption{(a-b) The global representation manifold $r_c$ for planetary orbits with coloring according to $r_n$ and $e$. (c-d) The state space of planetary orbits motion, with coloring according to $r$ and $\theta$.}
\label{fig:orbits2}
\end{figure}

We generate image sequences by varying $r_0$, $v_0^r$ and $v_0^{\theta}$ in the range $[1.5,2]$, $[0,0.2]$ and $[0.7,0.8]$ respectively and for each sequence, we generate $101$ image frames. If our model can learn correct dynamics, it is expected that the dynamic latent parameters $z$ lie in a two dimensional manifold correlated to the ground-truth states $(r, \theta)$, while the global representation $r_c$ of each sequence should also lie within a two dimensional manifold corresponding to the ground-truth system parameters $(r_n,e)$. 

\begin{figure}[htbp]
\centerline{\includegraphics[width=0.95\columnwidth]{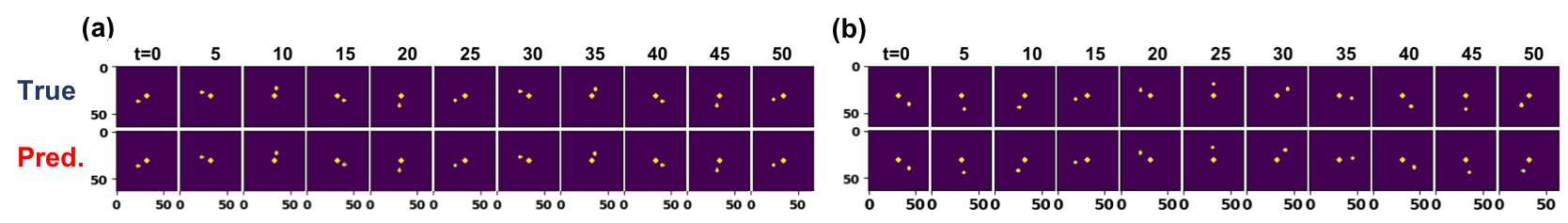}}
\caption{Two samples of image sequences of orbit motions in meta-test set for 50-step predictions. The first row shows the ground-truth images, while the second row shows the predicted ones.}
\label{fig:orbits4}
\end{figure}

In Figure \ref{fig:orbits2}(a)(b), we can clearly see that the learned global latent representation indeed lies in a two dimensional manifold. This is an \textit{amazing} result, as we take a long analytical effect in previous paragraphs to identify that the independent number of ground-truth system parameters which affect the dynamic evolution is two (i.e. $r_n$ and $e$) rather than three (i.e. $r_n$, $e$ and $\theta_n$). We again fit the latent representations $r_c$ with global parameters $r_n$ and $e$ using linear and quadratic regressions shown in Table \ref{tb1}. The fitted $R^2$ scores are very high which means a good match with the ground-truth global parameters. The learned dynamic state space in Figure \ref{fig:orbits2}(c)(d) also shows our model can infer the physically meaningful state representations. In Figure \ref{fig:orbits4} we also show two sets of multi-step predictions (50 time-steps) on the meta-test set, with both ground-truth (the first row) and predicted (the second row) images. It can be seen that the dynamic model we learned can make correct long-term prediction of the orbit motion. We also show the model performance together with $D=1$ in Table \ref{tb2}, the lower KL term values for $D=5$ also indicates that the overshooting technique is of great help to long-term predictions.

\section{Related Work}

Our work was inspired by the work of Garnelo et al. \cite{garnelo_neural_2018} applying NP to function regression and optimization. Kim et al. \cite{kim_attentive_2019} addressed the underfitting issue by incorporating attention mechanism. Eslami et al. \cite{eslami_neural_2018} and Kumar et al. \cite{kumar_consistent_2018} applied it to the scene representation and Singh et al. \cite{NIPS2019_9214} extended it to sequential cases. 

For the state transition modelling, Karl et al. \cite{karl_deep_2016} proposed Deep Variational Bayes Filters (DVBF), a linear Gaussian state space model (LGSSM) using linear Gaussian models to learn and to identify Markovian state space. Deisenroth et al. \cite{deisenroth_pilco:_2011} introduced PILCO, a data-efficient policy search method in model-based reinforcement learning, which uses GPs to model state transition dynamics explicitly incorporating model uncertainty into long-term planning, achieving unprecedented learning efficiency on high-dimensional control tasks. Watter et al. \cite{WatterSBR15} proposed Embed to Control (E2C) that learns a local linear dynamical transition model from raw pixel images which achieves promising results on a variety of complex control problems. Recently, Hafner et al. \cite{hafner_learning_2019} proposed Deep Planning Network (PlaNet), a model-based reinforcement learning method that learns the state dynamics from images and plans in latent space and achieves performance comparable to model-free algorithms in continuous control tasks.

For physical dynamics learning, Iten et al. \cite{iten_discovering_2018} used NP to discover physical concepts, but they did not learn the underlying Markovian state and dynamics, while the inputs to the model are low-dimensional states. Breen et al. \cite{breen_newton_2019} applied deep neural networks to solve the motion of chaotic three-body problem and achieved accurate solutions several orders faster than a state-of-the-art solver.
\section{Conclusions}

In this work, we propose a novel network architecture NeurPhy that can learn physical dynamics directly from image sequences. NeurPhy can not only correctly extract the global system parameters from the sequence and dynamic state from each frame, but also succeeds in predicting dynamic evolution in unseen cases. We apply our model for the characterization and prediction of damped pendulum and planetary orbits motion, and achieve promising results. Our architecture is quite general, which can be easily extended to model-based reinforcement learning, which we will explore in the future.

\section{Supplementary Material}

\subsection{Derivation of the lower bound objective function}

Now we derive the lower bound objective function of NeurPhy. For overshooting length $d$, Conditioned on context points $x_c$, the log-likelihood of data can be written as:
\begin{align}
&\ln p_d(x_{1:T}|x_c)\nonumber\\
&=\ln \int \prod_{t=1}^Tp(x_t|z_t)p(z_t|z_{t-d},r_c)p(r_c|x_c)dz_{1:T}dr_c\nonumber\\
&=\ln \int \prod_{t=1}^T\frac{q(z_t|x_{\leq t})}{q(z_t|x_{\leq t})}p(x_t|z_t)p(z_t|z_{t-d},r_c)p(r_c|x_c)dz_{1:T}dr_c.
\end{align}
Here, we use the Markovian state space asumption\cite{doerr_probabilistic_2018}, i.e., the dynamic state $z_t$ contains all the information needed for image $x_t$ reconstruction. Also, we assume the approximate posterior $q(z_t)$ does not depend on context $r_c$ explicitly given $x_{\leq t}$. From Jensen's inequality, we can get:
\begin{align}
&\ln \int \prod_{t=1}^T\frac{q(z_t|x_{\leq t})}{q(z_t|x_{\leq t})}p(x_t|z_t)p(z_t|z_{t-d},r_c)p(r_c|x_c)dz_{1:T}dr_c\nonumber\\
&\geq \mathbb{E}_{q(z_{1:T}|x_{\leq T})p(r_c|x_c)}\sum_{t=1}^T\left[\ln p(x_t|z_t)+\ln p(z_t|z_{t-d},r_c)-\ln q(z_t|x_{\leq t})\right]\nonumber\\
&= \mathbb{E}_{q(z_{1:T}|x_{\leq T})p(r_c|x_c)}\sum_{t=1}^T\left[\ln p(x_t|z_t)+\ln \mathbb{E}_{p(z_{t-1}|z_{t-d},r_c)}p(z_t|z_{t-1},r_c)-\ln q(z_t|x_{\leq t})\right]\nonumber\\
&\geq \mathbb{E}_{q(z_{1:T}|x_{\leq T})p(r_c|x_c)}\sum_{t=1}^T\left[\ln p(x_t|z_t)+\mathbb{E}_{p(z_{t-1}|z_{t-d},r_c)}\ln p(z_t|z_{t-1},r_c)-\ln q(z_t|x_{\leq t})\right]\nonumber\\
&= \sum_{t=1}^T \mathbb{E}_{q(z_t|x_{\leq t})}\ln p(x_t|z_t)-\mathop{\mathbb{E}\mathop{\text{KL}}(q(z_t|x_{\leq t})||p(z_t|z_{t-1},r_c))}_{p(z_{t-1}|z_{t-d},r_c)q(z_{t-d}|x_{\leq {t-d}})p(r_c|x_c)},
\end{align}
which is eq.(2) in the main paper.

\subsection{Derivation of the planetary orbits motion using the law of universal gravitation}

\begin{figure}[htbp]
\centerline{\includegraphics[width=0.5\columnwidth]{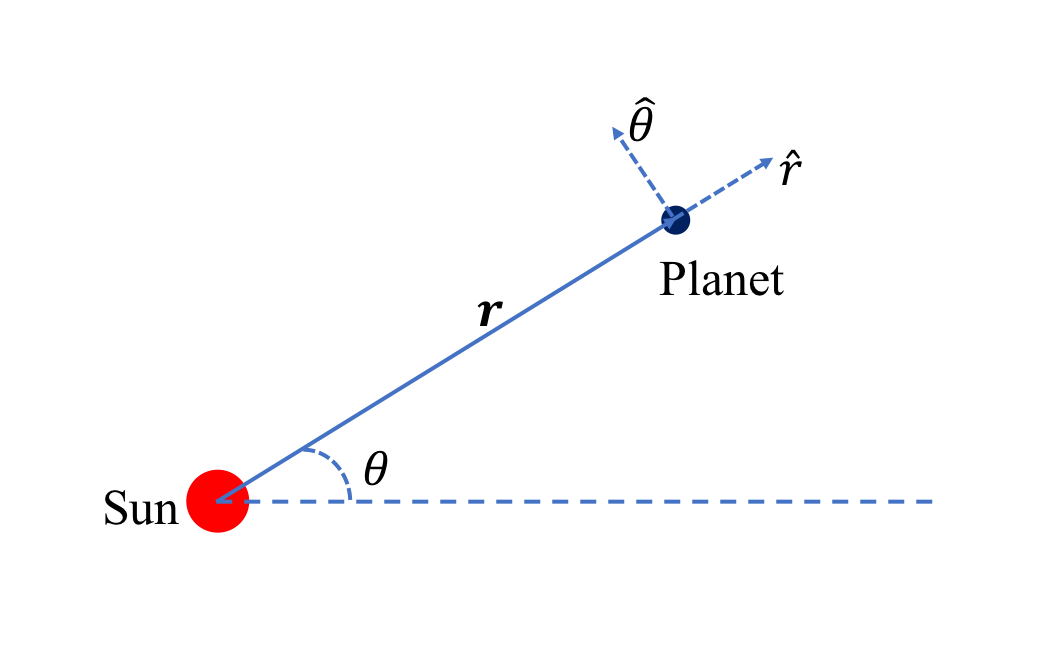}}
\caption{Schematic diagram of the planetary orbits motion with
r denoting the radius from planet to the central sun, $\theta$ denoting the angle between r and reference axis. $\hat{r}$ and $\hat{\theta}$ denote unit vectors along radial and angular direction respectively.}
\label{fig:aorbits2}
\end{figure}

\subsubsection{The elliptical orbits}

In this subsection, we derive the elliptical orbits motion obeyed by planets which circle around a central sun. We follow the derivation in reference \cite{roy_orbital_2004,cite_s1}. Assuming the mass of central sun $M$ is several orders larger than the mass of planet $m$, then the position of the central sun is approximately still and we only need to consider the motion of the planet. Using polar coordinates $(r,\theta)$, the position vector $\mathbf{r}$ can be written as (see Figure \ref{fig:aorbits2}):
\begin{equation}
\mathbf{r} = r\hat{r}.
\end{equation}
Here, symbols in bold font denote vectors, symbols with hat denotes unit vectors along the corresponding coordinate direction. Applying time derivative to $\mathbf{r}$, we can get:

\begin{align}
&\dot{\mathbf{r}} = \dot{r}\hat{r} + r\dot{\hat{r}} = \dot{r}\hat{r} + r\omega\hat{\theta},\\
&\ddot{\mathbf{r}} = (\ddot{r}-r\omega^2)\hat{r} + (2\dot{r}\omega+r\dot{\omega})\hat{\theta}.
\end{align}
Here, we rewrite angular velocity $\dot{\theta}$ as $\omega$ and have used identities $\hat{r} = \omega \hat{\theta}, \hat{\theta} = -\omega \hat{r}$ in the derivation.

Now, From the second law of motion $\mathbf{F} = m\mathbf{a}=m\ddot{\mathbf{r}}$ and the law of universal gravitation:
\begin{equation}
\mathbf{F} = -G\frac{Mm}{r^2}\hat{r},
\end{equation}
we have:
\begin{align}
&\ddot{r}-r\omega^2 = -\frac{GM}{r^2},\label{s7}\\
&2\dot{r}\omega+r\dot{\omega} = 0.\label{s8}
\end{align}
Here, $G$ denotes the gravitational constant and $\ddot{}$ denotes the second derivative of time. Eq. (\ref{s8}) can be rewritten to a total differential form:
\begin{align}
&2\dot{r}\omega+r\dot{\omega} = \frac{d}{dt}(r^2\omega) = 0\nonumber\\
&\Rightarrow r^2\omega = h,\label{s9}
\end{align}
where $h$ is a constant, which is proportional to angular momentum $l = mrv_\theta=mr^2\omega=mh$. Eq. (\ref{s9}) is also equivalent to the Kepler's second law of planetary motion.

Let $u=1/r$, we have:
\begin{align}
&\dot{r}= -\frac{\dot{u}}{u^2}=-\frac{\dot{\theta}}{u^2}\frac{du}{d\theta}=-h\frac{du}{d\theta},\nonumber\\
&\ddot{r}=-h\dot{\theta}\frac{d^2u}{d\theta^2}=-h^2u^2\frac{d^2u}{d\theta^2}.
\end{align}
Together with eq.(\ref{s7}), we can get:
\begin{align}
&\frac{d^2u}{d\theta^2} + u =\frac{GM}{h^2}.
\end{align}
By solving above equation, we have following general solution:
\begin{align}
&u =A\cos(\theta-\theta_n) + \frac{GM}{h^2}.
\end{align}

So the planetary orbit $r(\theta)$ is:
\begin{align}
&r(\theta) =\frac{1}{A\cos(\theta-\theta_n) + \frac{GM}{h^2}}.
\end{align}
Here, $\theta_n$ denotes the angle of the major axis and $A$ is a coefficient.

The nearest radius called the perihelion distance $r_n$ can be calculated by setting $\theta = \theta_n$, i.e.,
\begin{align}
&r_n =\frac{1}{A + \frac{GM}{h^2}}.
\end{align}
Define the eccentricity $e=Ah^2/GM$, we have:
\begin{align}
&r_n =\frac{h^2}{GM(1+e)}.\label{s15}
\end{align}

Then we can rewrite $r$ using $r_n$ as:
\begin{align}
r = r_n\frac{1+e}{1+e\cos(\theta-\theta_n)}.\label{s16}
\end{align}
This is the eq. (6) used in the main paper.

\subsubsection{Derive system global parameters from initial condition}

Now we derive the global parameters $(h,e,\theta_n,r_n)$ from the initial condition $(r_0,v_0^r,v_0^\theta)$. Note that $h$ is a constant, we have:
\begin{align}
h = r_0v_0^\theta.\label{s17}
\end{align}
From the law of energy conservation, we have:
\begin{align}
\frac{1}{2}m((v_0^r)^2+(v_0^\theta)^2) - \frac{GMm}{r_0} = \frac{1}{2}m(v_n^\theta)^2 - \frac{GMm}{r_n}.\label{s18}
\end{align}
Here, the left-hand side is the energy at initial moment, while the right-hand side represents the energy at perihelion $r_n$. Together with $h = r_nv_n^\theta$ and eq. (\ref{s15}), we have:
\begin{align}
\frac{G^2M^2}{2h^2}(e^2-1) = \frac{1}{2}((v_0^r)^2+(v_0^\theta)^2) - \frac{GM}{r_0}.
\end{align}

From the above equation, we can get the eccentricity $e$, and from eq. (\ref{s16}), we have:
\begin{align}
r_0 = r_n\frac{1+e}{1+e\cos(\theta_0-\theta_n)} = r_n\frac{1+e}{1+e\cos(\theta_n)},\label{s20}
\end{align}
from which we can get $\theta_n$. Here, we assume the initial angle $\theta_0=0$.

So, from eq. (\ref{s15})(\ref{s17})(\ref{s18})(\ref{s20}), we can see that the set $(r_0,v_0^r,v_0^\theta)$ has one-to-one correspondence with $(r_n,e,\theta_n)$, as claimed in the main paper.

\subsubsection{Derive the dynamic transition equations}

From Eq.(\ref{s9}), we have:
\begin{align}
\dot{\theta} = \frac{h}{r^2(\theta)}.\label{s21}
\end{align}

Integrate above equation with time $t$, we can derive the time evolution of $\theta$. And from the relation of $\theta$ with $r$ (eq. (\ref{s16})), we can also get the time evolution of $r$. 

The differential equation eq. (\ref{s21}) is a transcendental equation, which has no analytical solution. In our setting, we only have to derive the state transition equations from $(r_{t},\theta_{t})$ to $(r_{t+1},\theta_{t+1})$. From eq. (\ref{s21}), we have:
\begin{align}
&\theta_{t+1} = \theta_{t} + \Delta t\cdot\frac{h}{r_t^2},
\end{align}

and from eq. (\ref{s16}), we have:

\begin{align}
&r_{t} = r_n\frac{1+e}{1+e\cos(\theta_t-\theta_n)},\\
&r_{t+1} = r_n\frac{1+e}{1+e\cos(\theta_{t+1}-\theta_n)}.
\end{align}

These transition equations correspond to eq. (7) in the main paper.

\subsection{Experimental results for low-dimensional inputs using Neurphy}
Our model is quite general, if inputs are not raw images but low-dimensional variables (such as positions of pendulum end-point), our model can also be applied to the systems by changing all convolutional layers in NeurPhy to multilayer perceptrons. As we will show below, the learning tasks actually become easier.

\subsubsection{Damped pendulum}
\begin{figure}[htbp]
\centerline{\includegraphics[width=0.8\columnwidth]{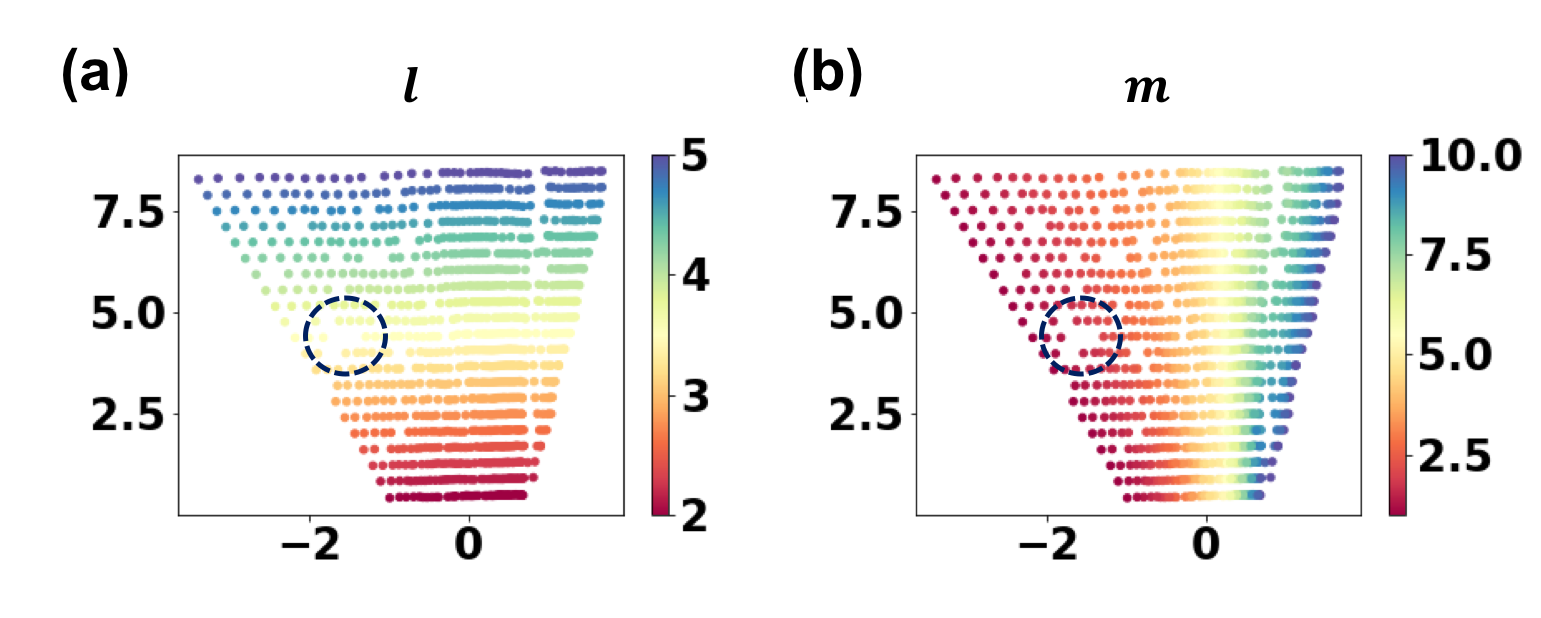}}
\caption{The global representation manifold for damped pendulum with colormaps denoting underlying (a) $l$ and (b) $m$. The hole marked by dashed circles represent samples in meta-test.}
\label{fig:ps1}
\end{figure}
\begin{figure}[htbp]
\centerline{\includegraphics[width=0.8\columnwidth]{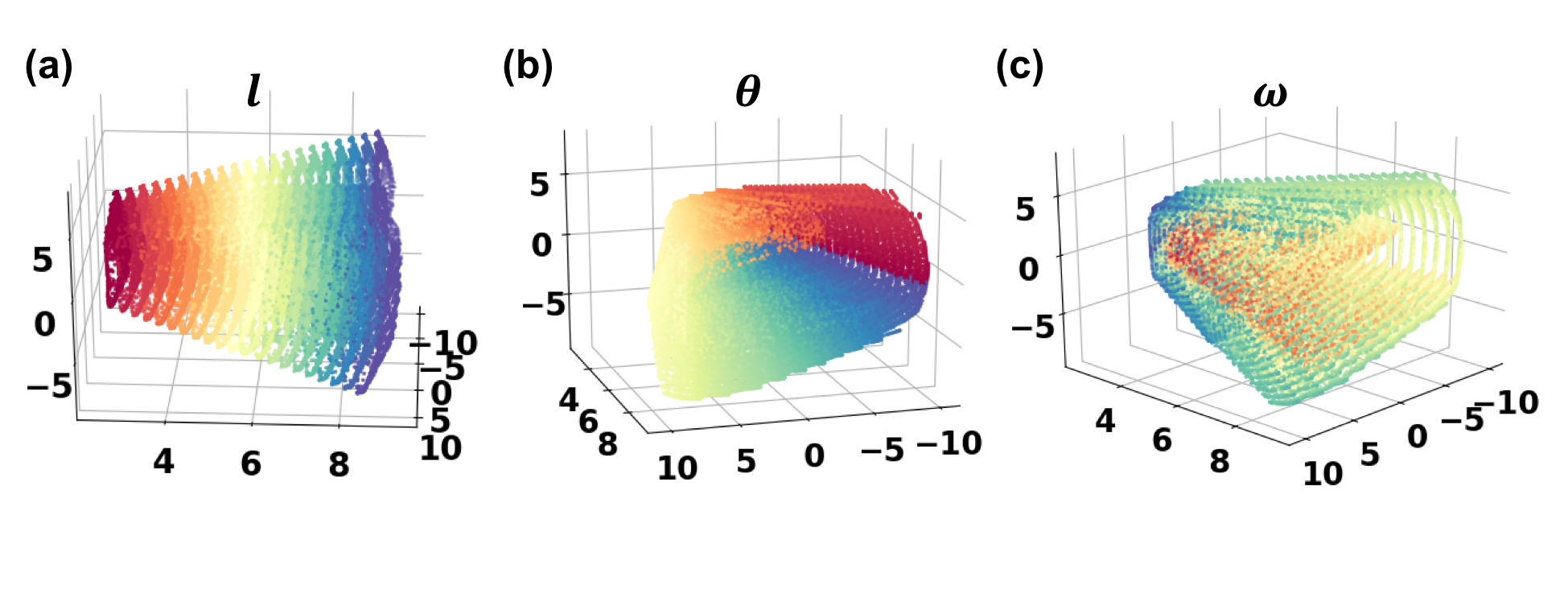}}
\caption{The dynamic state space for damped pendulum with colormaps denoting underlying (a) $l$, (b) $\theta$ and (c) $\omega$.}
\label{fig:ps2}
\end{figure}

We use the position coordinates $(x,y)$ of pendulum end-points as inputs (i.e., the images of dimension $64\times 64\times 1$ in the main paper are replaced by positions of dimension $2$), following the same routine as in the main paper, we run experiments with various global parameters $l,m$ ($l\in[2,5]$ and $m\in[2,10]$).

The learned global representations are shown in Figure \ref{fig:ps1}. In Figure \ref{fig:ps1}(a) and (b), colormaps denote the ground-truth $l$ and $m$ respectively. We can see a clear correspondence between the learned manifold and the true global parameters. Furthermore, there are some holes between data points in the figure (such as the places marked by dashed circles), corresponding to the meta-test samples. In Figure \ref{fig:ps2}, we plot the dynamic state space of damped pendulum with colormaps denoting the ground-truth dynamic state $(l,\theta,\omega)$. The learned state space lies in a 3-dimensional manifold, which shows good correspondences with underlying dynamic states.

\begin{figure}[htbp]
\centerline{\includegraphics[width=0.8\columnwidth]{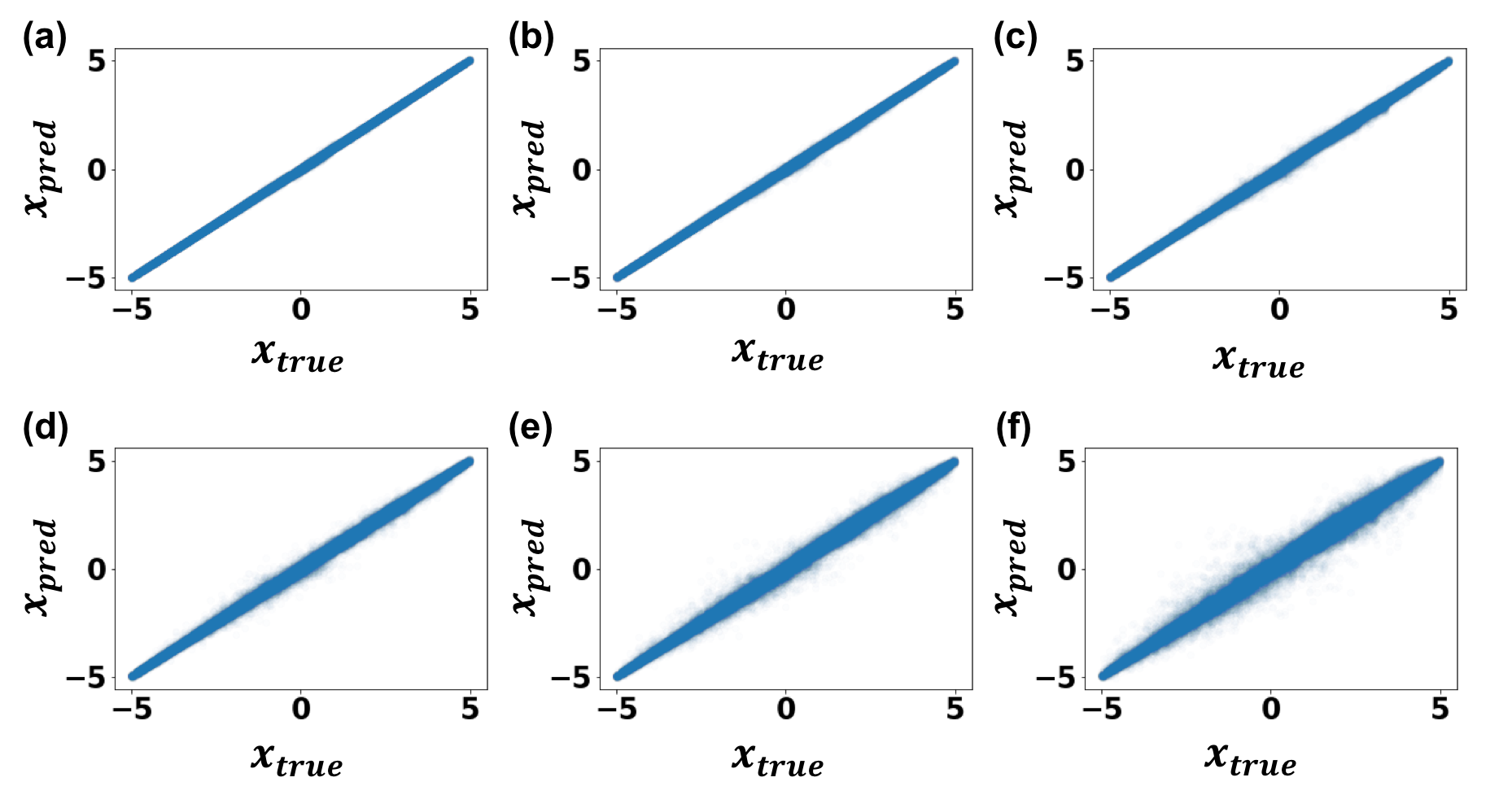}}
\caption{The true and predicted x coordinates of pendulum end-points with different overshooting lengths d: (a) $d = 0$, (b) $d=1$, (c) $d=2$, (d) $d=3$, (e) $d=4$ and (f) $d=5$.}
\label{fig:ps3}
\end{figure}

To show the performance of our model on low-dimensional inputs, we plot the x coordinates of pendulum end-points for training set in Figure \ref{fig:ps3}, with x axes denoting the true values and y axes denoting predicted ones. We can see a perfect match between the true and the predicted pendulum's position. When the overshooting length d is longer, the prediction get worse. In Table 3, we calculate the mean square errors (MSEs) of the true and the predicted pendulum's position for training, test and meta-test (20 and 2 context samples) sets. $T+d$ represents the result with overshooting length $d$, in the table, d changes from $0$ to $5$. We can see that on both training and test sets, MSEs are quite small, which get a bit larger on the meta-test set.

\begin{table}[htbp]
 \center
\begin{tabular}{|c|c|c|c|c|c|c|}
\hline
 \textit{MSE}&  \textbf{T+0}&  \textbf{T+1}&  \textbf{T+2}& \textbf{T+3}&\textbf{T+4}&\textbf{T+5} \\
 \hline
 \textbf{Training}&  0.00082&  0.0033&  0.010& 0.025&0.062&0.14   \\
 \hline
 \textbf{Test}&  0.00095&  0.0040&  0.014& 0.035&0.086&0.19  \\
 \hline
 \textbf{Meta-test (20 contexts)}&  0.00081&  0.011&  0.046& 0.14&0.28&0.56\\
 \hline
 \textbf{Meta-test (2 contexts)}&  0.00082&  0.016&  0.064& 0.19&0.36&0.72\\
 \hline
\end{tabular}
\caption{MSE for damped pendulum with different overshooting lengths.} 
\end{table}

\begin{figure}[htb]
\centerline{\includegraphics[width=0.8\columnwidth]{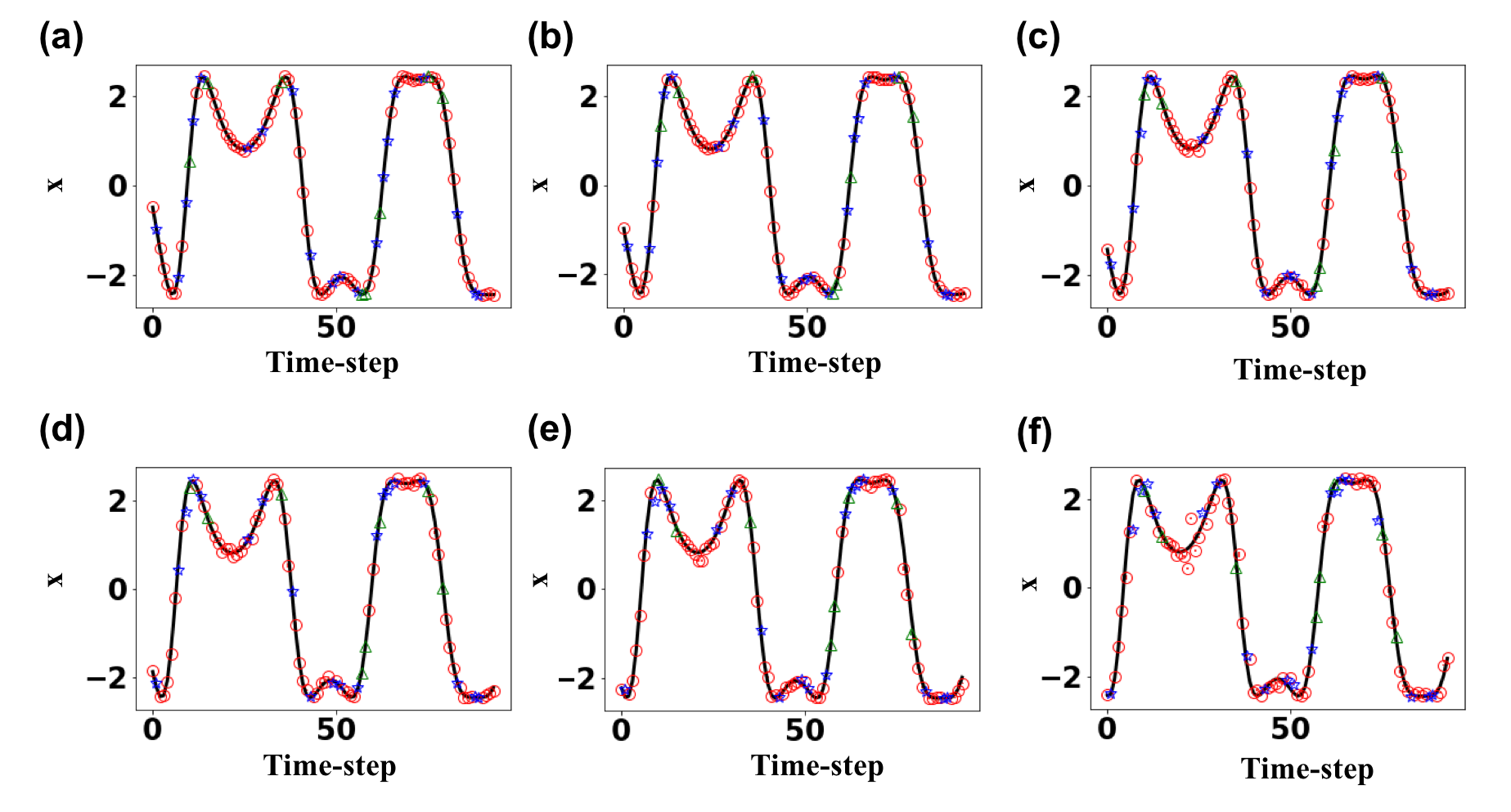}}
\caption{The true and predicted x coordinates of damped pendulum for a training sequence. The solid lines denote true x coordinates, the blue stars are context points, the red circles are target points and the green triangles are test points. The x-axis denotes time step and we plot results for different overshooting lengths in (a) $d=0$, (b) $d=1$, (c) $d=2$, (d) $d=3$, (e) $d=4$ and (f) $d= 5$.}
\label{fig:ps4}
\end{figure}

\begin{figure}[htb]
\centerline{\includegraphics[width=0.8\columnwidth]{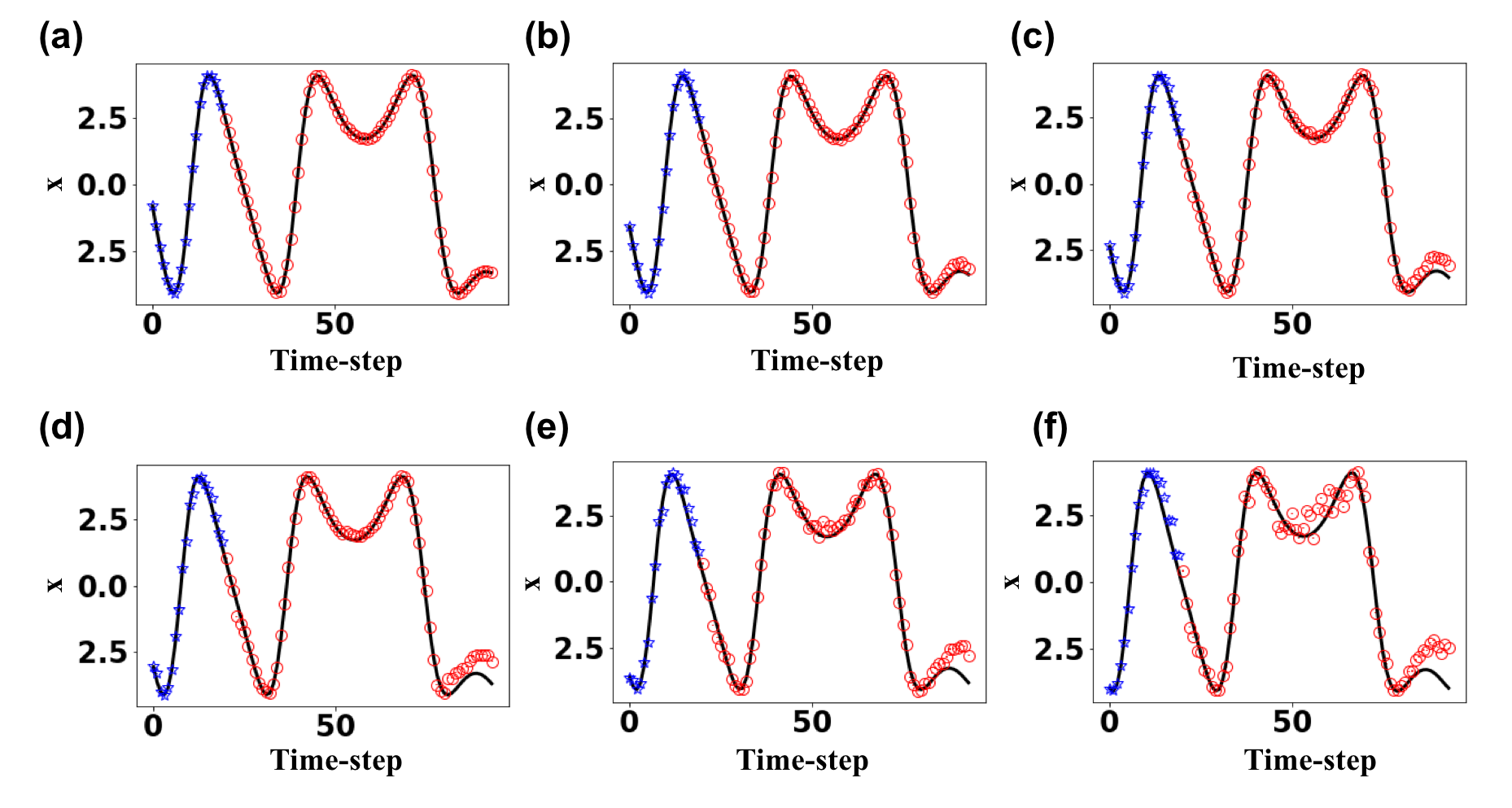}}
\caption{The true and predicted x coordinates of damped pendulum for a meta-test sequence with 20 context samples. The solid lines denote true x coordinates, the blue stars are context points, the red circles are test points for prediction. The x-axis denotes time step and we plot results for different overshooting lengths in (a) $d=0$, (b) $d=1$, (c) $d=2$, (d) $d=3$, (e) $d=4$ and (f) $d= 5$.}
\label{fig:ps5}
\end{figure}

\begin{figure}[htb]
\centerline{\includegraphics[width=0.8\columnwidth]{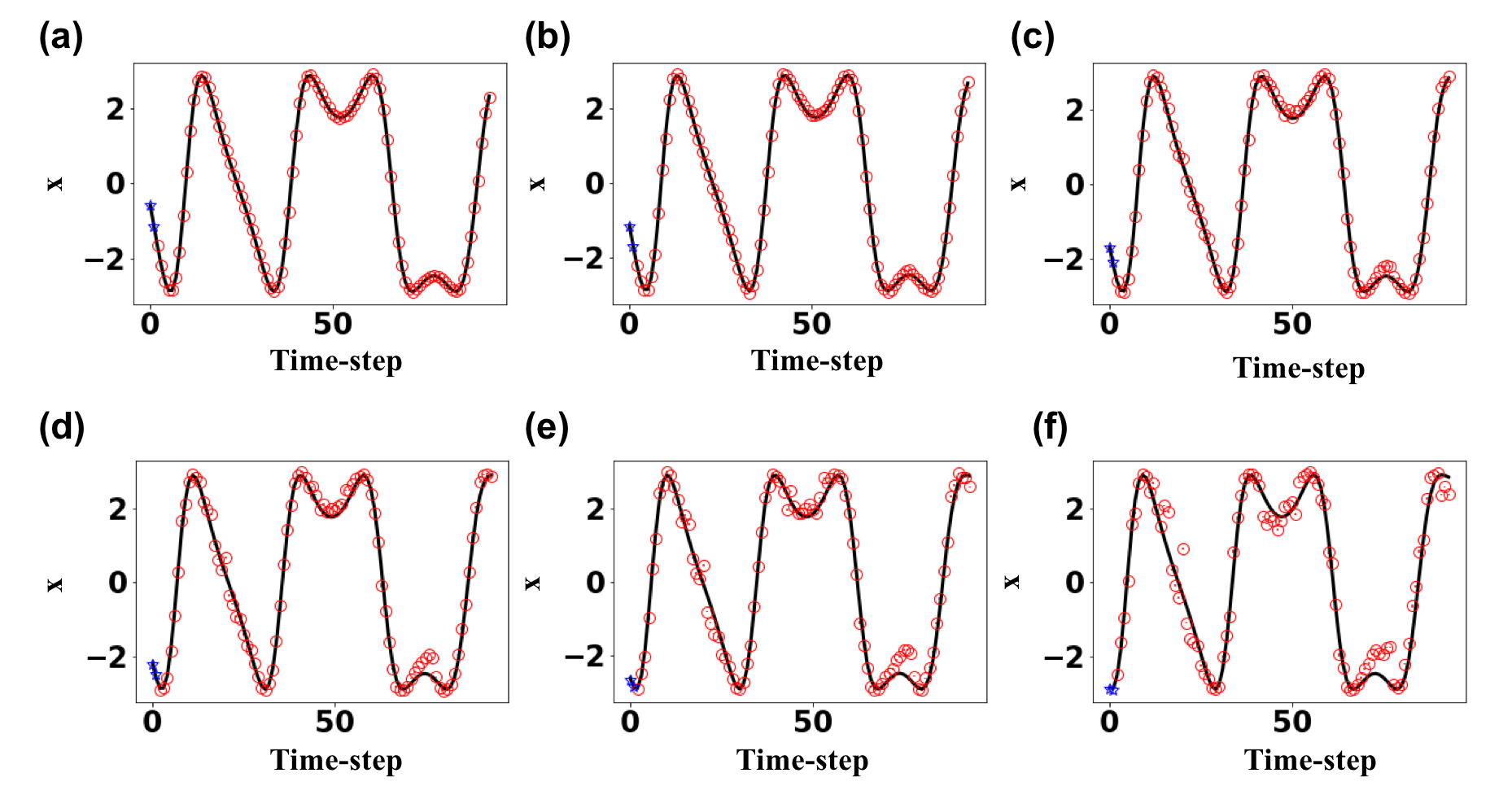}}
\caption{The true and predicted x coordinates of damped pendulum for a meta-test sequence with 2 context samples. The solid lines denote true x coordinates, the blue stars are context points, the red circles are test points for prediction. The x-axis denotes time step and we plot results for different overshooting lengths in (a) $d=0$, (b) $d=1$, (c) $d=2$, (d) $d=3$, (e) $d=4$ and (f) $d= 5$.}
\label{fig:ps6}
\end{figure}

In Figure \ref{fig:ps4}, we plot the x coordinates of one training sequence with different predicting length (overshooting length d). The solid lines are ground-truth coordinates of the sequence and points in symbols are predicted ones for both context, target and test samples. We can clearly see that our model fits well. In Figure \ref{fig:ps5} and \ref{fig:ps6}, we apply our model to one meta-test sample whose underlying global parameters $(l,m)$ is unseen in the training set. In Figure \ref{fig:ps5}, as the same as the training setting, we give the model 20 context samples, but in Figure \ref{fig:ps6}, we only give the model 2 context samples to see if our meta-learning algorithm can deal with fewer context samples. We can see that our model can still make quite good prediction even only 2 context samples are given, this can also be seen in the last row of Table 3, the MSEs for different overshooting lengths is only slightly larger than those given 20 context samples.


\subsubsection{Planetary orbits motion}

\begin{figure}[htbp]
\centerline{\includegraphics[width=0.8\columnwidth]{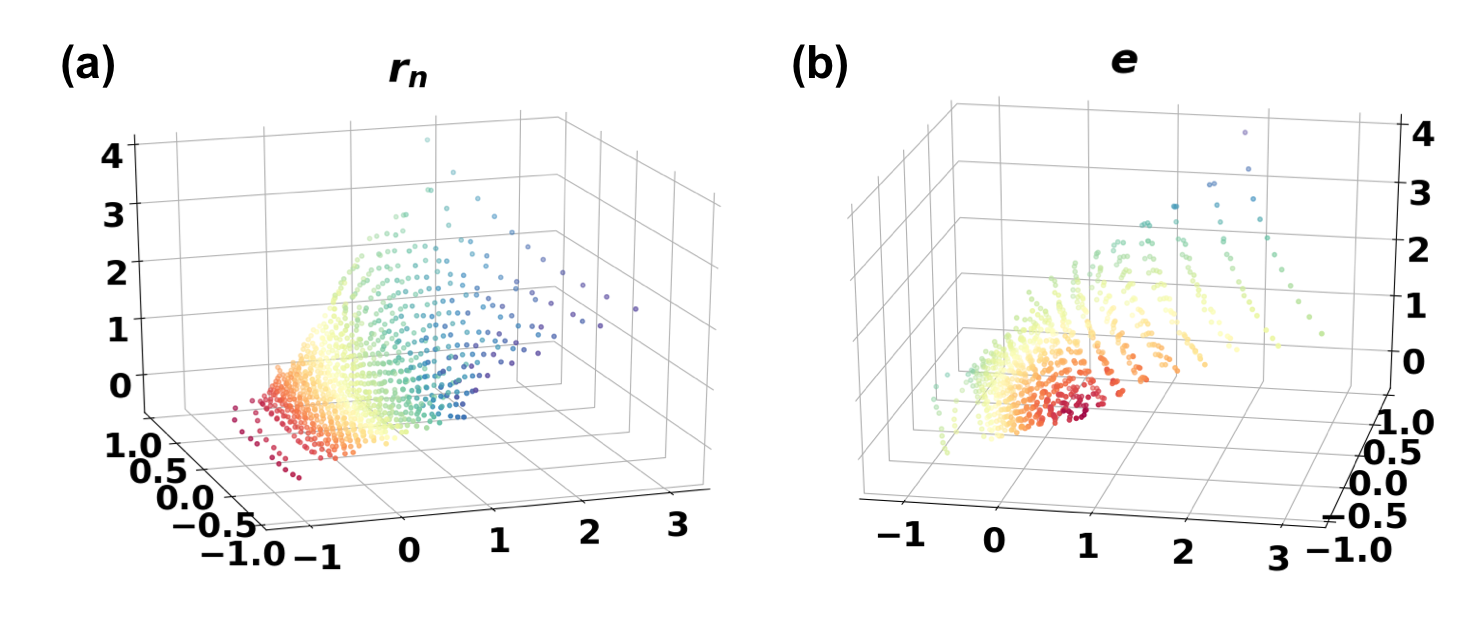}}
\caption{The global representation manifold $r$ for planetary orbits with colormaps denoting (a) $r_n$ and (b) $e$.}
\label{fig:zorbits1}
\end{figure}

\begin{figure}[htbp]
\centerline{\includegraphics[width=0.8\columnwidth]{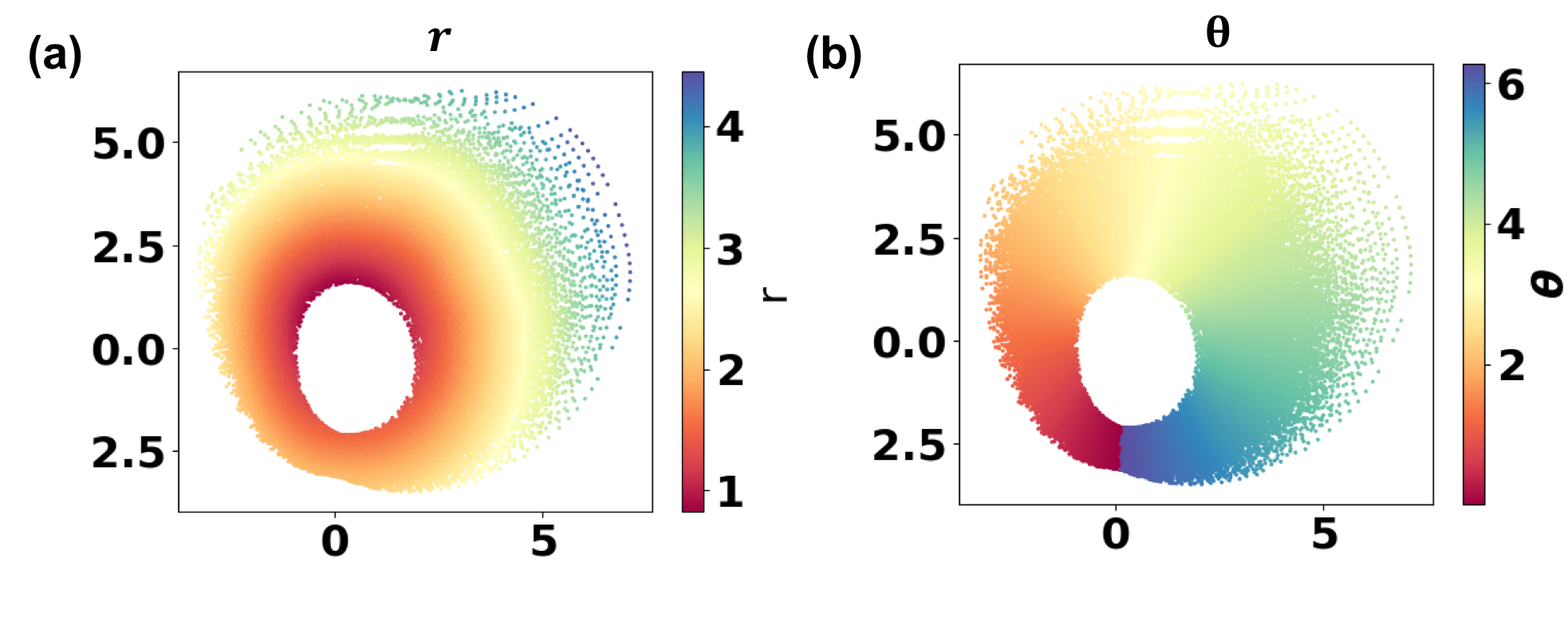}}
\caption{The dynamic state space for planetary orbits motion with colormaps denoting underlying (a) $r$, (b) $\theta$.}
\label{fig:zorbits2}
\end{figure}

\begin{figure}[htbp]
\centerline{\includegraphics[width=0.8\columnwidth]{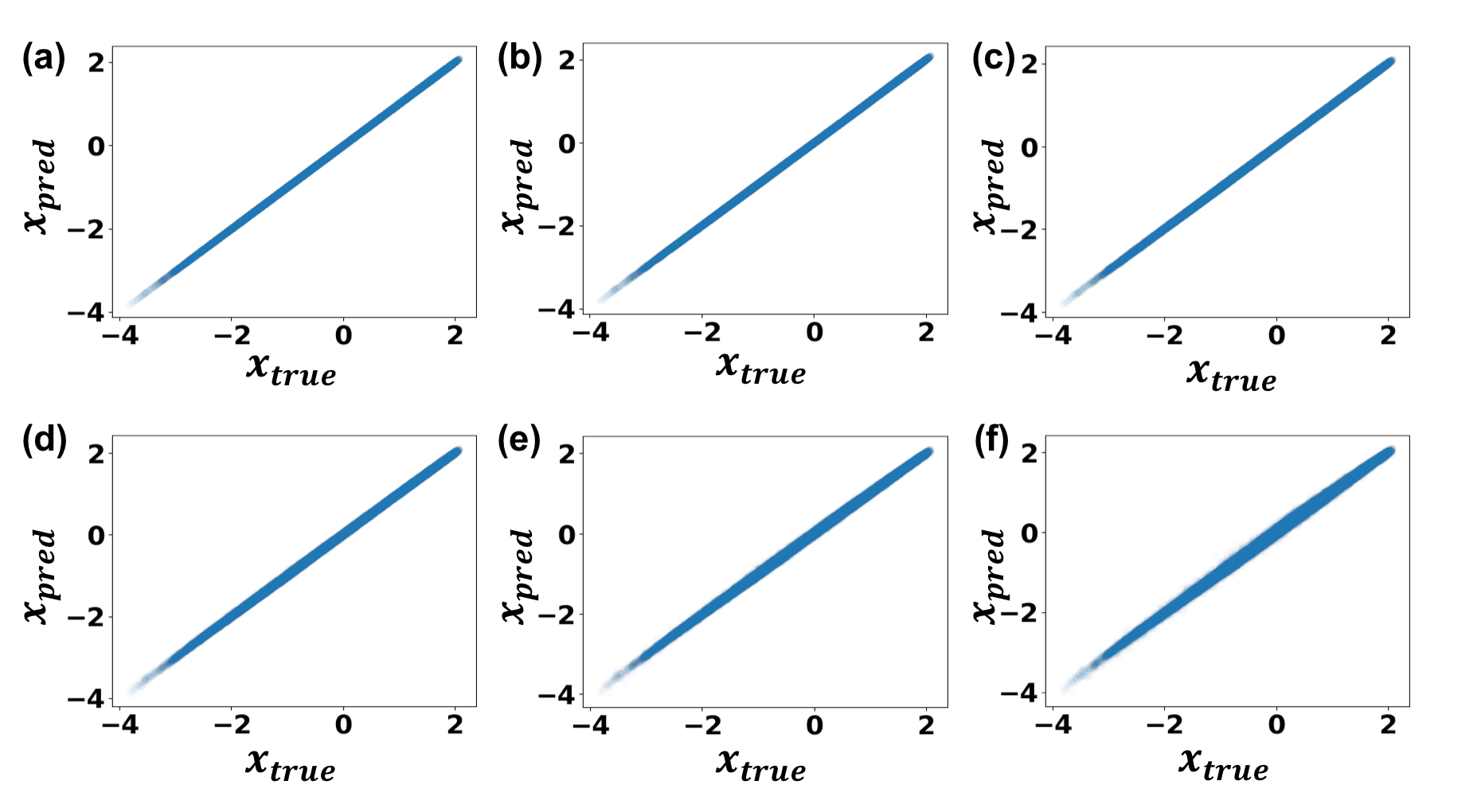}}
\caption{The true and predicted x coordinates of planet's positions with different overshooting lengths d: (a) $d = 0$, (b) $d=1$, (c) $d=2$, (d) $d=3$, (e) $d=4$ and (f) $d=5$.}
\label{fig:zorbits3}
\end{figure}

\begin{figure}[htbp]
\centerline{\includegraphics[width=0.8\columnwidth]{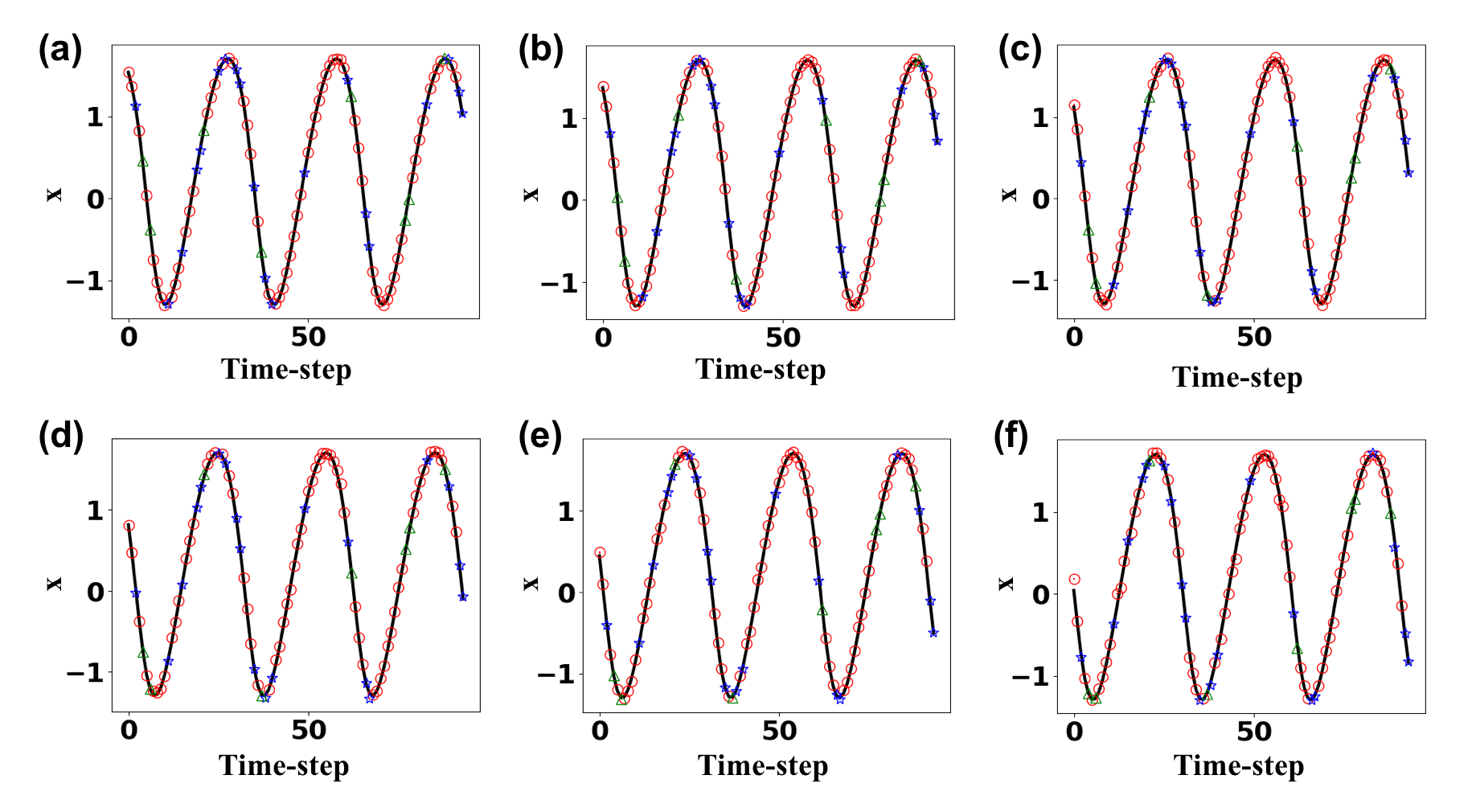}}
\caption{The true and predicted x coordinates of planet's positions for a training sequence. The solid lines denote true x coordinates, the blue stars are context points, the red circles are target points and the green triangles are test points. The x-axis denotes time step and we plot results for different overshooting lengths in (a) $d=0$, (b) $d=1$, (c) $d=2$, (d) $d=3$, (e) $d=4$ and (f) $d= 5$.}
\label{fig:zorbits4}
\end{figure}

\begin{figure}[htbp]
\centerline{\includegraphics[width=0.8\columnwidth]{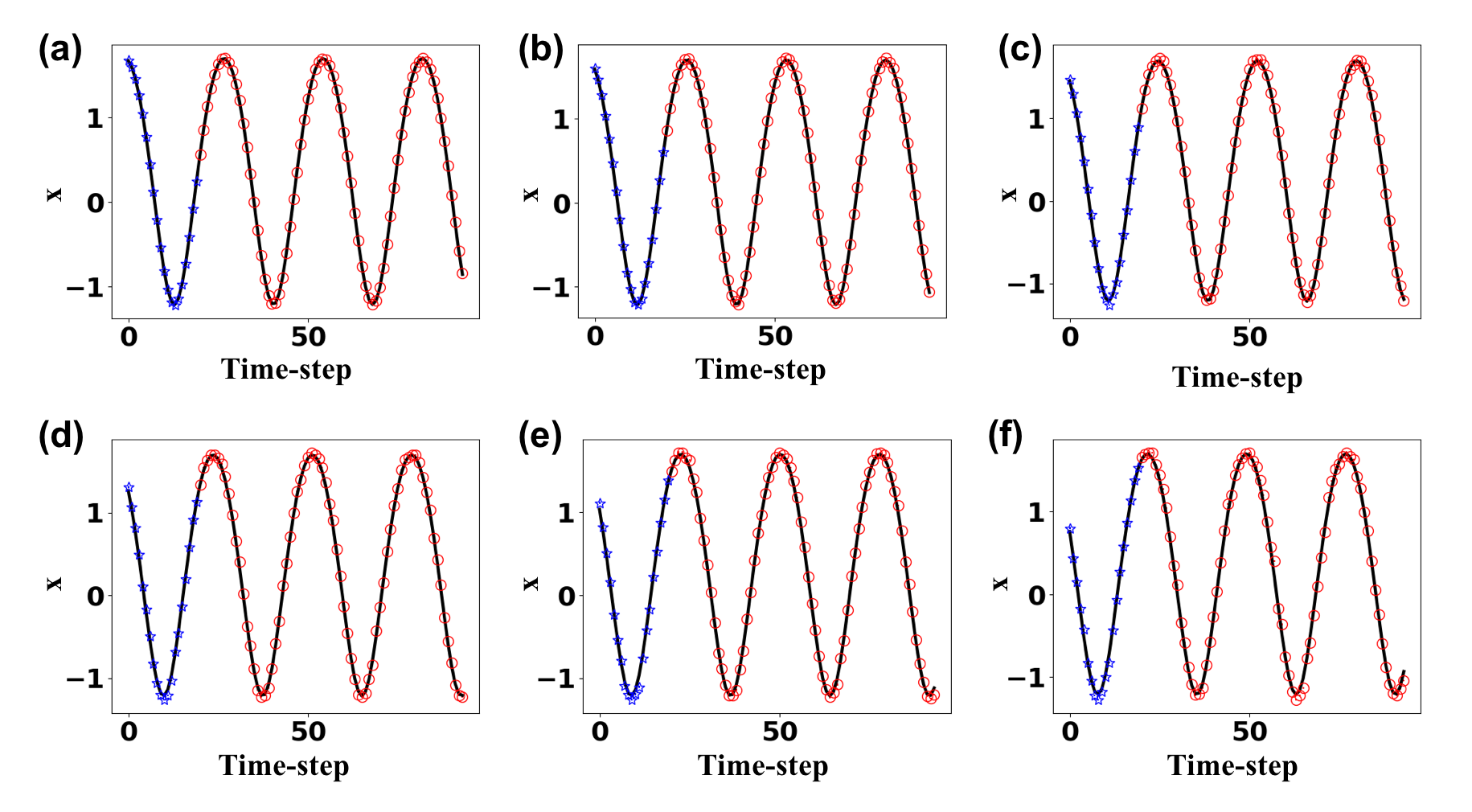}}
\caption{The true and predicted x coordinates of planet's positions for a meta-test sequence with 20 context samples. The solid lines denote true x coordinates, the blue stars are context points, the red circles are test points for prediction. The x-axis denotes time step and we plot results for different overshooting lengths in (a) $d=0$, (b) $d=1$, (c) $d=2$, (d) $d=3$, (e) $d=4$ and (f) $d= 5$.}
\label{fig:zorbits5}
\end{figure}

\begin{figure}[htbp]
\centerline{\includegraphics[width=0.8\columnwidth]{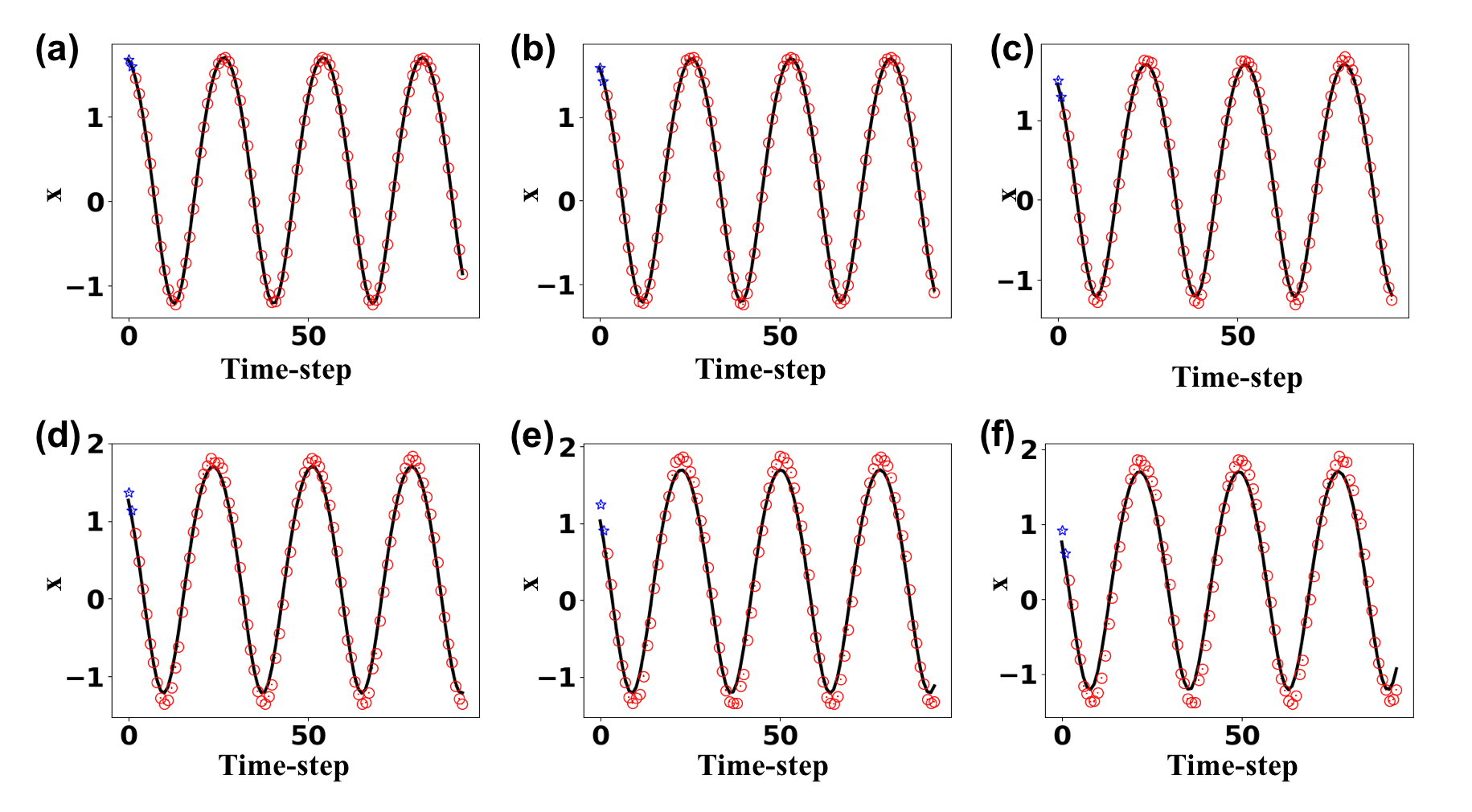}}
\caption{The true and predicted x coordinates of planet's positions for a meta-test sequence with 2 context samples. The solid lines denote true x coordinates, the blue stars are context points, the red circles are test points for prediction. The x-axis denotes time step and we plot results for different overshooting lengths in (a) $d=0$, (b) $d=1$, (c) $d=2$, (d) $d=3$, (e) $d=4$ and (f) $d= 5$.}
\label{fig:zorbits6}
\end{figure}

We also apply the low-dimensional inputs model for the prediction of planetary orbits motion. All the settings are the same as in the main paper except the inputs are changed to planet's position coordinates $(x,y)$. In Figure \ref{fig:zorbits1} and \ref{fig:zorbits2}, we plot the learned global representation and dynamic state space. In Figure \ref{fig:zorbits3}, we plot the true and predicted x coordinates of planet's positions. In Table 4, we give the MSEs of predictions. In Figure \ref{fig:zorbits4}, \ref{fig:zorbits5} and \ref{fig:zorbits6}, we plot the performance for a training and meta-test (both 20 and 2 context samples) sequence. We can see that our model indeed can learn the underlying global and local representations together with the dynamic transition process. Also note that it is easier for our model to learn the \textit{planetary orbits motion} than the \textit{damped pendulum}.

\begin{table}[htbp]
\center
\begin{tabular}{|c|c|c|c|c|c|c|}
\hline
 \textit{MSE}&  \textbf{T+0}&  \textbf{T+1}&  \textbf{T+2}& \textbf{T+3}&\textbf{T+4}&\textbf{T+5} \\
 \hline
 \textbf{Training}&  0.000067&  0.00023&  0.00053& 0.00094&0.0016&0.0034   \\
 \hline
 \textbf{Test}&  0.000068&  0.00023&  0.00055& 0.00097&0.0017&0.0034   \\
 \hline
 \textbf{Meta-test (20 contexts)}&  0.000068&  0.0019&  0.013& 0.050&0.11&0.16 \\
 \hline
 \textbf{Meta-test (2 contexts)}&  0.000067&  0.0041&  0.030& 0.20&0.11&0.30 \\
 \hline
\end{tabular}
\caption{MSE for planetary orbits motion with different overshooting lengths.} 
\end{table}


\subsection{Network structure of NeurPhy}
\subsubsection{For images inputs}

For \textit{Global representation model} and \textit{Recognition model}, we use convolutional neural networks, each with four layers with filter sizes of [32,64,64,128], kernel sizes of $4\times 4$, strides of 2 and ReLU activations. For \textit{Observation model}, we use deconvolutional neural networks with filter sizes of [64,64,32,1], kernel sizes of [5,5,6,6], strides of 2 and three layers of ReLU activations, and the activation of last layer is Sigmoid. For \textit{State space model}, we just use a 4-layer forward neural network with cell units [128,128,64,16] and ReLU activations. The last layer is further connected to a dense layer with units size $2\times dim_z$, where $dim_z$ is the dimension of the latent dynamic state, which is set it to 3 in our experiments. First $dim_z$ units generates the mean value of $z$ and the second $dim_z$ units produces the standard derivation.

\subsubsection{For low-dimensional variables inputs}

The network architecture is almost the same as the case for the images, except that the convolutional layers are placed by multi-layers forward networks. For \textit{Global representation model}, it is replaced by a 4-layer forward neural network with cell units of [128,128,64,16] and ReLU activations. For \textit{Recognition model}, we only use a 2-layer forward neural network with cell units of [32,16] and ReLU activations. For \textit{damped pendulum} system, we set $\beta_d = 1$ for all $d \in [1,D]$, batch size $B=2$ and run 1000 epochs. For \textit{planetary orbits motion}, we set $\beta_d=1$ for all $d \in [1,D]$, batch size $B=5$ and run 500 epochs.

{\small
\bibliography{nn_physics_v7_arxiv.bbl}
}

\end{document}